%% file: main.tex
\documentclass[11pt,a4paper]{article}
\usepackage[margin=1in]{geometry}
\usepackage[T1]{fontenc}
\usepackage[utf8]{inputenc}
\usepackage{newtxtext,newtxmath}
\usepackage{amsmath,booktabs,array,graphicx}
\usepackage[protrusion=false]{microtype}
\usepackage{placeins,adjustbox,float}
\usepackage[font=small,labelfont=bf]{caption}
\usepackage{cite}
\usepackage[hidelinks]{hyperref}

\newcommand{\pp}{\,\mathrm{pp}}

\let\origus\_\renewcommand{\_}{\origus\allowbreak}
\title{How Much Do Teacher Bridges Matter in Interleaved On-Policy Distillation?}
\author{Leran Chen \and Zhiyuan Yao \and Lingnan Kong}
\date{}
\hypersetup{pdftitle={How Much Do Teacher Bridges Matter in Interleaved On-Policy Distillation?},pdfauthor={Leran Chen, Zhiyuan Yao, Lingnan Kong}}
\makeatletter
\renewcommand{\maketitle}{%
\begin{center}
{\fontsize{20}{24}\selectfont\bfseries\@title\par}
\vspace{18pt}
\noindent \begin{minipage}[t]{0.315\textwidth}\centering
{\large\bfseries Leran Chen\par}\vspace{4pt}
{\fontsize{9.5}{11.5}\selectfont\parbox[t][25pt][t]{\linewidth}{\centering The University of Sydney}}\par
{\fontsize{8.5}{10}\selectfont\texttt{\href{mailto:lche0711@uni.sydney.edu.au}{lche0711@uni.sydney.edu.au}}}
\end{minipage}
\hfill
\begin{minipage}[t]{0.315\textwidth}\centering
{\large\bfseries Zhiyuan Yao\par}\vspace{4pt}
{\fontsize{9.5}{11.5}\selectfont\parbox[t][25pt][t]{\linewidth}{\centering University of California\\San Diego}}\par
{\fontsize{8.5}{10}\selectfont\texttt{\href{mailto:c8yao@ucsd.edu}{c8yao@ucsd.edu}}}
\end{minipage}
\hfill
\begin{minipage}[t]{0.315\textwidth}\centering
{\large\bfseries Lingnan Kong\par}\vspace{4pt}
{\fontsize{9.5}{11.5}\selectfont\parbox[t][25pt][t]{\linewidth}{\centering The University of Sydney}}\par
{\fontsize{8.5}{10}\selectfont\texttt{\href{mailto:lkon0802@uni.sydney.edu.au}{lkon0802@uni.sydney.edu.au}}}
\end{minipage}
\end{center}\vspace{12pt}}
\makeatother
\input{macros_auto}

\input{macros_v4_auto}
\input{macros_v5_auto}
\input{macros_cond_auto}
\input{macros_v7_auto}

\begin{document}
\maketitle

\begin{abstract}
On-policy distillation (OPD) trains a student using teacher predictions on student-generated responses. Some methods insert teacher-written segments, or bridges, but disagree on how to train on those tokens. We compare no loss, forward KL and reverse KL on identical trajectories using Qwen3-1.7B as student and Qwen3-4B-Instruct-2507 as teacher. We separate gradients by which model wrote each token. Under our short training runs and BF16 inference, pre-specified tests detected no added accuracy benefit from teacher text as context or as a training target. A one-paragraph bridge after the student's first paragraph is only \mTokShareB{} of the tokens and contributes a small gradient nearly orthogonal to the student-token gradient. In this main comparison, the three loss treatments gave similar gradient directions and were statistically equivalent in accuracy within a $\pm5$ percentage-point margin. An exploratory pilot placing the bridge where failed responses disagreed most with the teacher suggested a benefit from teacher-token loss. A pre-specified replication did not confirm this, and fourfold weighting showed no significant accuracy benefit. Two limitations bound these findings. Among the tested amounts of training on fixed responses, gains peaked after four updates before the student increasingly failed to stop generating. None of the tested online training variants met the criteria for stable stopping and improved accuracy. Our evaluations used BF16 inference, which rounds away most of the small weight changes. When evaluated in exact FP32, the tested replication models frequently reached the response-length limit. Our accuracy findings therefore concern BF16 inference; they do not establish that teacher-token supervision is generally unnecessary.
\end{abstract}

\section{Introduction}
On-policy distillation (OPD) trains a student on its own samples under dense token-level teacher supervision. Several recent variants let the teacher write part of a trajectory: as a prefix (TCOD, ReOPD), as interleaved turns or segments (Guided-OPD, DASH-OPD, Relay-OPD), or as a validated bridge (FutureBridge-OPD). A teacher-written segment can matter in two distinct ways. It changes the \emph{states} the student is subsequently supervised on, and it can itself be a \emph{target} of direct supervision. Existing methods make different choices about the second role: no loss on teacher tokens (TCOD-B2F, ReOPD), forward KL on teacher tokens (Guided-OPD), or several teacher-leg objectives (Relay-OPD). Prior work already includes loss-direction and teacher-loss ablations~\cite{guided,uopd} and component-gradient measurements~\cite{segopd}. We compare teacher-token loss treatments on identical interleaved trajectories and measure author-separated gradient norms, cosines and changes in gradient direction.

We study a minimal interleaved setting, $S\!\to\!T\!\to\!S$. A student prefix $A$ is followed by one teacher paragraph $B$ and the student's continuation $C$. On \emph{the same} fixed trajectories we compare reverse KL on all positions, no loss on $B$ (with $B$ kept as context), forward KL on $B$, a random mask that matches the kept-token count and terminal-EOS state, and pure-student trajectories that share the same prefix. We make five contributions:
(i) an author-segmented measurement of where the supervision signal lives: token share, KL-mass share, gradient-norm share, inter-segment gradient cosine and one-step cross-segment interference;
(ii) a pre-specified, multi-seed controlled comparison whose central prediction is derived from (i) and tested by equivalence; (iii) a second pre-specified experiment that tests, with a dose scan, the exploratory signal the first one produced; (iv) a pilot that moves the teacher to the point where the student's failed answers derail and varies the teacher text, its placement and teacher-token supervision, holding the realized text fixed for the loss comparison (Section~\ref{sec:cond}); and (v) a pre-specified replication of that pilot with a weight dose on the teacher term, together with an exact-FP32 check of the evaluation numerics (Section~\ref{sec:v7}).
At the derail point the teacher segment is longer and carries a larger, aligned share of the gradient, and in the pilot the loss on its tokens appeared to matter. The replication did not confirm this, and a fourfold weight on the teacher term, which turned the gradient by \rvAngleW{} degrees, showed no significant accuracy benefit. Across both placements we detected no benefit from the states a teacher segment induces or from the loss on its tokens. The FP32 check adds a caveat to every result: under the BF16 inference used throughout, most of a small update is rounded away, and the tested replication models no longer terminate reliably in full precision after 4 updates.
Because a termination collapse capped every dose we could study, we add an exploratory diagnosis of that collapse under 100 steps of online training (Section~\ref{sec:online}). It probes stopping preferences at student end states, checks their sensitivity to the chat template, and tests four repairs and a length-matched teacher.
We do not propose a new algorithm. The context-only use of teacher tokens and error-triggered teacher intervention both have direct precedents, and we cite them as such.

\section{Setting and objective}
For a fixed response $y_i$ of length $L_i$ with states $s_{it}$, student $p_\theta$ and frozen teacher $q$, every arm optimizes
\begin{equation}
\begin{aligned}
\mathcal L(\theta)&=\frac1N\sum_{i=1}^N\frac1{L_i}\sum_{t=1}^{L_i} w_{it}\,D_{it}(\theta),\\[3pt]
D_{it}&\in\Big\{\textstyle\sum_v p_\theta\log\frac{p_\theta}{q},\ \sum_v q\log\frac{q}{p_\theta}\Big\},
\end{aligned}
\end{equation}
with full-vocabulary divergences. Masks never remove tokens from later contexts, and they never renormalize the denominator. Segment gradients $g_A,g_B,g_C$ are the gradients of the segment-restricted sums. By linearity, an arm's gradient is the sum of its segment gradients. We compute every segment gradient with the same validated operator and verify that the sum matches the one-shot gradient within the GPU nondeterminism floor.

\section{Experimental design}
The fixed-trajectory main study is labelled S3; S4 denotes the single-seed, fresh-rollout on-policy check (Section~\ref{sec:onpolicy}).
\paragraph{Models and trajectories.} Student Qwen3-1.7B (non-thinking), frozen teacher Qwen3-4B-Instruct-2507, one RTX PRO 6000. For each of three seeds, 128 DeepMath training questions receive a student prefix up to its first paragraph boundary (cap 256). The teacher then writes one paragraph from the prefix (cap 1024), and the student continues to EOS or a 4096-token total cap. The same prefix also receives a pure-student continuation.
\paragraph{Arms.} allRKL, AC (no loss on $B$), mixedKL (forward KL on $B$), randomK and pureS. Each arm restarts from the initial student and takes 4 AdamW updates of 32 trajectories each, i.e.\ one pass over the seed's 128 trajectories (lr $3\times10^{-6}$, clip 1). Within a seed, all arms use the same batch permutation.
\paragraph{Evaluation.} The student answers alone, without the teacher, using Qwen3's recommended non-thinking sampling (T$=0.7$, top-$k$ 20, top-$p$ 0.8). Common random numbers are used: each question has its own random stream, shared by all models. E1: the first 256 of 512 frozen, previously unevaluated in-domain DeepMath questions (primary; the size was reduced from 512 before any arm comparison, Section~\ref{sec:deviations}). E2: 256 GSM8K test questions (transfer; a standard boxed-answer instruction is appended). Strict accuracy requires an explicit final answer that matches the gold answer and ends with EOS. All models keep FP32 weights and run under BF16 autocast.
\paragraph{A dose-dependent termination collapse (exploratory, motivated the final dose).} The first version of the study used 16 updates of 8 trajectories. After this dose, the student stopped terminating. On 64 historical questions, 42--43 responses hit the 4096-token cap under both greedy decoding and the recommended sampling, against 3 for the initial student. Most truncated responses were endless self-corrections (``Wait, no. Wait: \dots''), not exact loops. An earlier screen with 4 updates at the same learning rate kept termination intact. We therefore paused before any arm comparison and reduced the dose to 4 updates. We report this collapse as an observation, not as a tested effect (Section~\ref{sec:collapse}).
\paragraph{Advance specification.} We use \emph{pre-specified} for hypotheses and analysis rules documented internally before the relevant comparisons. The plans were frozen in local files with SHA-256 hashes; they were not publicly registered (Appendix~\ref{app:repro}). The original protocol, data hashes and analysis plan preceded the main run (\texttt{protocol\_v2.txt}). The main-study amendments (\texttt{protocol\_v3\_amendment.txt}) changed the decoding, dose and E1 size after nontermination had been observed in the initial trained arm, but before between-arm accuracy comparisons (Section~\ref{sec:deviations}). The earlier greedy evaluations were retained as exploratory records and excluded from the final arm comparisons. A pilot on previously used trajectories fixed the mechanistic prediction, and a pre-written rule excluded a projection arm.

\section{Where does the supervision live?}
\label{sec:mechanism}
\paragraph{Measurements.} For each update, a segment set $S\in\{A,B,C\}$ and a divergence direction give (i) the KL mass $\mathcal L_S$, i.e.\ the segment's share of the arm's normalized loss, (ii) the segment-gradient norm $\|g_S\|$, and (iii) cosines between segment gradients. These are computed at the current parameters of every arm and every step, so they are measured, not extrapolated. We report $\cos(g_B,g_{AC})$ for reverse and forward KL on $B$. We also report the one-step cross-segment interference: the change in the $A{+}C$ loss on the same batch after the actual update.

\paragraph{Pilot (exploratory; previously used trajectories).} Before the main study, we measured these quantities at the initial parameters on the 32 teacher-assisted trajectories of an earlier screen. These trajectories had already been inspected, so this is exploratory evidence. The teacher bridge was 3.9\% of response tokens (1{,}438 of 36{,}468). It carried 2--4\% of the KL mass. Its gradient norm was $0.75$--$1.02$, against $6.4$--$7.9$ for $g_{AC}$ (median per-trajectory ratio $0.075$). Batch-level $\cos(g_{B}^{\mathrm{rev}},g_{AC})$ ranged from $+0.07$ to $+0.15$, and $\cos(g_{B}^{\mathrm{fwd}},g_{AC})$ from $-0.04$ to $+0.06$ (median $+0.046$). Across individual trajectories, 31--38\% of cosines were negative, but all of them were small in magnitude. The pre-clip gradient norm was $\approx7.6$, so every update was clipped to norm 1. A rule written before the pilot added a projection (PCGrad-style) arm only if the median $\cos(g_B^{\mathrm{fwd}},g_{AC})$ was negative. It was not, so the arm was not run.

\paragraph{Prediction.} The teacher bridge contributes a small, nearly orthogonal component, and the gradient is rescaled by clipping. Treating $B$ differently should therefore barely change the gradient direction. This is a mechanistic prediction, not an identity for AdamW parameter steps. We predicted that allRKL, AC and mixedKL would be equivalent within $\pm5\pp$ and tested this by TOST in the main study.

\begin{figure}[!htbp]\centering
\includegraphics[width=\linewidth]{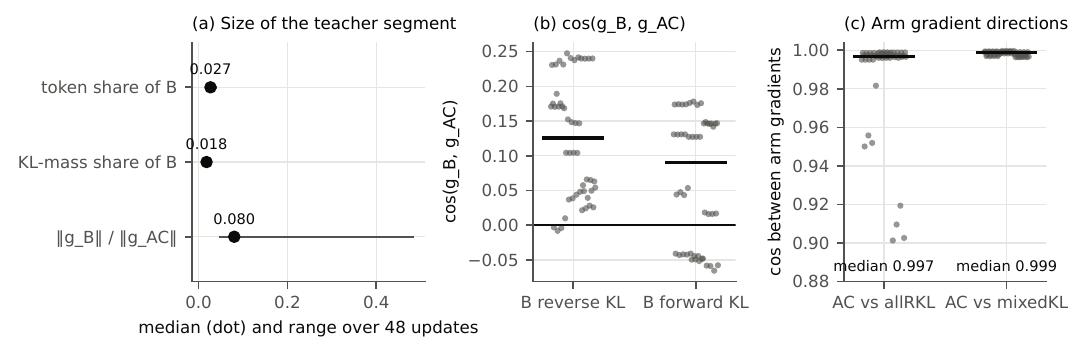}
\caption{Author-segmented gradient geometry over all \mNupd{} updates of the teacher-assisted arms. (a) Token share, KL-mass share and relative gradient norm of the teacher bridge $B$. (b) Cosine between $g_B$ and the student-segment gradient $g_{AC}$. (c) Cosine between the gradients of two arms at identical parameters.}\label{fig:mech}
\end{figure}

\paragraph{Main study.} Across all \mNupd{} updates of the four arms trained on teacher-assisted trajectories (3 seeds $\times$ 4 updates), the teacher bridge made up a median \mTokShareB{} of response tokens (range \mTokShareBLo--\mTokShareBHi). It carried only \mKLShareB{} of the reverse-KL mass, less than its token share. Its gradient norm was a median \mNormRatioB{} of $\|g_{AC}\|$ (range \mNormRatioBLo--\mNormRatioBHi). Median $\cos(g_B,g_{AC})$ was \mCosBrevAC{} for reverse KL (range \mCosBrevACLo{} to \mCosBrevACHi) and \mCosBfwdAC{} for forward KL. The bridge gradient therefore neither dominates nor conflicts with the student-segment gradient. Linearity gives the exact cosine between the gradients of two arms at the same parameters: $\cos(g_{AC},g_{AC}+g_B)$. Its median was \mCosUpdAll{} for AC vs.\ allRKL (minimum \mCosUpdAllLo) and \mCosUpdMixed{} for AC vs.\ mixedKL (minimum \mCosUpdMixedLo). Every gradient was clipped (median pre-clip norm \mGradNorm), so the optimizer received gradients of equal norm in almost the same direction; this does not imply equal AdamW parameter-step lengths (Figure~\ref{fig:mech}).

\section{Controlled comparison}
\label{sec:controlled}
\input{tables_auto_preprint}
Table~\ref{tab:e1} reports E1 accuracy for every arm and seed. Against the initial student, every trained arm gained 4.6--8.1\,pp, with all 95\% intervals excluding zero. The trained students also wrote about 28\% longer responses and ended with EOS slightly less often (Figure~\ref{fig:len}). On GSM8K (E2), all models stayed near the initial 87.5\%, with changes from $-1.6$ to $-0.8$\,pp and every interval including zero (Table~\ref{tab:e2}).

\paragraph{Comparison with a matched random perturbation.} The initial student is a single sampled run, while every arm is a mean over three trained models. We therefore also compare against a matched random perturbation (noiseAC), which moves $\theta_0$ by the AC update's per-tensor norm in a random direction. noiseAC alone scored $+2.5$\,pp above the initial student (95\% CI $[-2.0,+6.8]$), with initial-like response lengths. Against noiseAC, the three teacher-bridge treatments gained only $+2.1$ to $+2.6$\,pp, and all intervals included zero (Table~\ref{tab:noise}). The random-mask arm gained $+3.6$\,pp $[+0.4,+7.2]$. Pure-student trajectories gained $+5.6$\,pp $[+2.2,+9.1]$, the largest gain over a no-learning perturbation of the same size.

\begin{figure}[!htbp]\centering
\includegraphics[width=\linewidth]{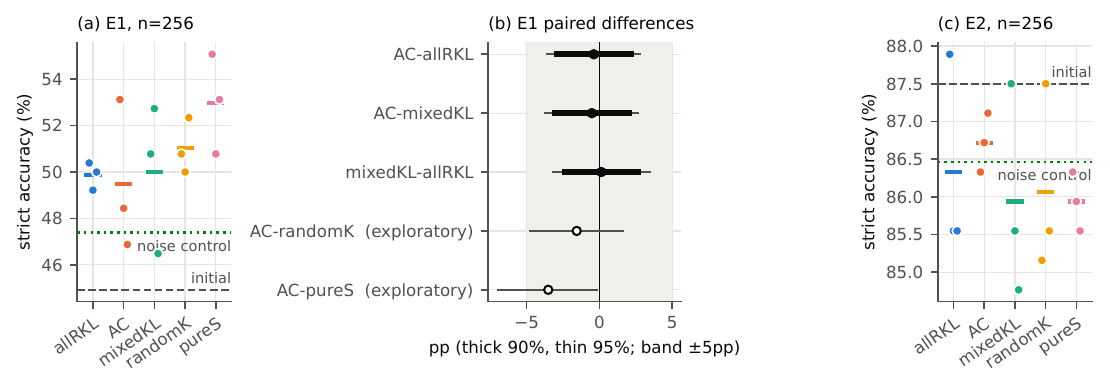}
\caption{(a, c) Strict accuracy per arm; dots are seeds, bars are means, and the dashed line is the initial student. (b) E1 paired differences with 90\% (thick) and 95\% (thin) intervals; the shaded band is the $\pm5$\,pp equivalence margin. Open markers are exploratory.}\label{fig:effects}
\end{figure}

\paragraph{Teacher-token treatment is immaterial here.} The three treatments of the teacher bridge were statistically equivalent within the pre-specified $\pm5$\,pp margin (Table~\ref{tab:contrasts}, Figure~\ref{fig:effects}): AC$-$allRKL $-0.4$\,pp, AC$-$mixedKL $-0.5$\,pp, mixedKL$-$allRKL $+0.1$\,pp. All 90\% intervals lie within $\pm3.3$\,pp. The confirmatory superiority family stopped at its first step (AC vs.\ allRKL, not significant), so no superiority claim is made for any later contrast.

\paragraph{No benefit from teacher-induced states (exploratory).} The random mask that matches token count and EOS state performed like the others (AC$-$randomK $-1.6$\,pp, 95\% CI $[-4.8,+1.7]$). Pure-student trajectories, which share the same prefix but contain no teacher paragraph, were the best arm in two of three seeds. AC$-$pureS was $-3.5$\,pp (95\% CI $[-7.0,-0.1]$, sign-flip $p=0.055$). Together with the noise comparison, this suggested that the continuation after a teacher bridge might be a \emph{less} useful training signal than the student's own continuation. Both contrasts lie outside the confirmatory family, and AC vs.\ pureS was the best of five arms, so we tested it separately (Section~\ref{sec:confirm}). It was not confirmed.

\section{On-policy check}
\label{sec:onpolicy}
In S4, allRKL, AC and mixedKL were trained truly on-policy, with one seed each. Each of the 4 updates first sampled 32 fresh $A|B|C$ trajectories from the \emph{current} student, using the same questions, schedule and loss. The geometry replicated exactly. The bridge was 2.8\% of tokens with a gradient-norm ratio of 0.076--0.077, $\cos(g_B,g_{AC})$ was 0.08--0.11, and the cosine between arm gradients was 0.997--0.999. E1 accuracy was 49.6\% (allRKL), 45.7\% (AC) and 47.3\% (mixedKL), against 44.9\% for the initial student; every interval against the initial student included zero. With a single seed the contrasts were too imprecise to establish equivalence: AC$-$allRKL was $-3.9$\,pp with 90\% interval $[-8.6,+0.8]$. In the fixed-trajectory study, the seed-to-seed spread of a single arm was of similar size (AC: 120--136 of 256). We therefore read S4 as confirming the mechanism under on-policy sampling, while saying nothing new about outcome differences.

\section{Confirmatory test: does the bridge help as a state?}
\label{sec:confirm}
\paragraph{Design.} A second protocol (\texttt{protocol\_v4.txt}) was frozen before its formal trained models were evaluated. It places AC$-$pureS first and alone: the primary test is E1 at 4 updates, two-sided $\alpha=0.05$. The same 3 seeds and trajectories are reused. Each arm is trained once for 8 updates of 32 trajectories (two passes, the second in a fresh fixed order) and evaluated at 2, 4 and 8 updates on the first 512 frozen E1 questions and on E2. Doses 2 and 8 are secondary (Holm). Everything else matches the main study.

\paragraph{Result.} The primary contrast was \vPrimMean\,pp (95\% CI $[\vPrimLo,\vPrimHi]$, $p=\vPrimP$), in the opposite direction from the exploratory $-3.5$\,pp, and not significant (Table~\ref{tab:v4contrast}). Neither secondary dose was significant. The interval excludes a pure-student advantage larger than 1.5\,pp at this dose. The exploratory signal of Section~\ref{sec:controlled} was most likely selection noise: it was the best of five arms on 256 questions.

\begin{figure}[!htbp]\centering
\includegraphics[width=\linewidth]{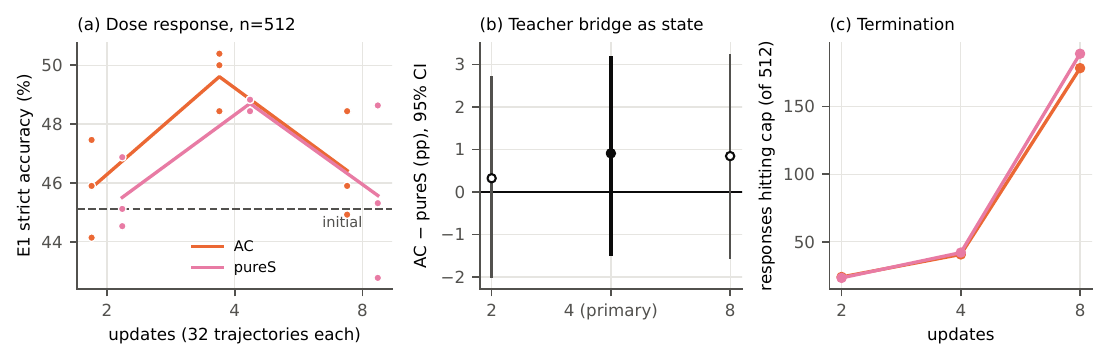}
\caption{Confirmatory dose scan. (a) E1 accuracy (dots: seeds; lines: means; dashed: initial student). (b) AC$-$pureS with 95\% intervals; filled marker: pre-specified primary test. (c) Mean E1 responses hitting the 4096-token cap.}\label{fig:dose}
\end{figure}

\paragraph{Dose response.} Both arms followed the same inverted U (Figure~\ref{fig:dose}, Table~\ref{tab:v4}). Neither gained at 2 updates. Both gained at 4 (AC \vACfourInit, pureS \vPSfourInit\,pp over the initial student, both intervals excluding zero; part of this may be perturbation, cf.\ Table~\ref{tab:noise}). At 8 updates termination collapsed: on average \vCapEight{} of 512 responses hit the cap, against \vCapFour{} at 4 updates and \vCapTwo{} at 2. The gain over the initial student vanished. The AC$-$pureS gap did not trend with dose (slope \vSlope\,pp per doubling).

\input{tables_v4_auto_preprint}

\section{Teacher intervention at the derail point (exploratory pilot)}
\label{sec:cond}
\paragraph{Question.} The bridge above follows the student's first paragraph whether or not the student goes wrong. Triggered methods use proxies for failure: with our model pair, Relay-OPD~\cite{relay} hands over when the teacher would start a reflection and the student would not, while MAESTRO~\cite{maestro} adapts intervention to paragraph-level disagreement. FutureBridge-OPD~\cite{futurebridge} validates a teacher bridge through paired future continuations; UOPD~\cite{uopd} uses teacher support for proposed student actions; Seg-OPD~\cite{segopd} selects uncertain segments from failed responses and trains a preference for teacher redrafts rather than inserting them. Prior work already ablates placement, execution or supervision, including UOPD's teacher-loss removal. We complement these controls by varying placement and continuation on the same failed questions and varying the loss on identical realized teacher text. The pilot's protocol, gate and analysis script were fixed before any trained model was evaluated (\texttt{protocol\_cond\_pilot.txt}).

\paragraph{Failures and the derail point.} The initial student answered \cvPool{} training questions used nowhere else in this study (T$=1$, cap 4096). \cvFail{} answers (\cvFailRate\%) were strictly wrong, \cvFailNoEos{} of them at the cap. For each wrong answer we computed the full-vocabulary reverse KL of every token under $\theta_0$. The \emph{derail point} is the paragraph with the highest mean KL among the eligible ones: not the first paragraph, starting before token 3072, at least 8 tokens long, and not after the first \texttt{\textbackslash boxed}. The \emph{control point} is a paragraph drawn by hash from the same set, at least two paragraphs away. Every failure had both. Over the \cvNtrainAll{} questions used for training, the derail point started at a median token \cvDerailStart{} (a median \cvDerailRel{} of the way through the answer), and only \cvDerailParaOne{} coincided with the main study's position. Its median paragraph KL was \cvDerailKL, against \cvControlKL{} at the control point. Each of three seeds received a disjoint block of \cvNtrain{} questions.

\paragraph{Arms.} \emph{condT}: the student prefix up to the derail point, a teacher segment, then the student's continuation to EOS or 4096 tokens in total, trained with the AC loss (no loss on the teacher segment). The teacher writes until the first paragraph boundary after 64 tokens, or until EOS (cap 384; median \cvTeachLen{} tokens, usually several paragraphs), longer than the main study's one-paragraph bridge. \emph{condS}: the same prefix continued by the student, with reverse KL on all positions as in pureS. \emph{randT}: condT at the control point. \emph{condT$_{\mathrm{KL}}$}: the condT trajectories with reverse KL on all positions, teacher segment included, as in allRKL. \emph{trunc}: the prefix alone. Optimizer, dose (4 updates of 32 from $\theta_0$), evaluation and statistics are those of the main study, on E1 only. The re-evaluated initial student matched the main study's initial evaluation on all 256 questions, response for response. Four contrasts were pre-specified: H1 condT$-$condS (teacher text as context), H2 condT$-$randT (placement), H3 condT$_{\mathrm{KL}}-$condT (loss on teacher tokens) and H4 condS$-$trunc (the continuation itself). The protocol predicted H1 and H2 to be positive and H3 to be an equivalence. A manipulation check before training asked whether condT trajectories rescue more failures than randT ones; it guides the reading of H2 and does not gate training. A full study would follow only if this check passed and H1 was at least $+2$\,pp.

\input{tables_cond_auto_preprint}

\begin{figure}[!htbp]\centering
\includegraphics[width=\linewidth]{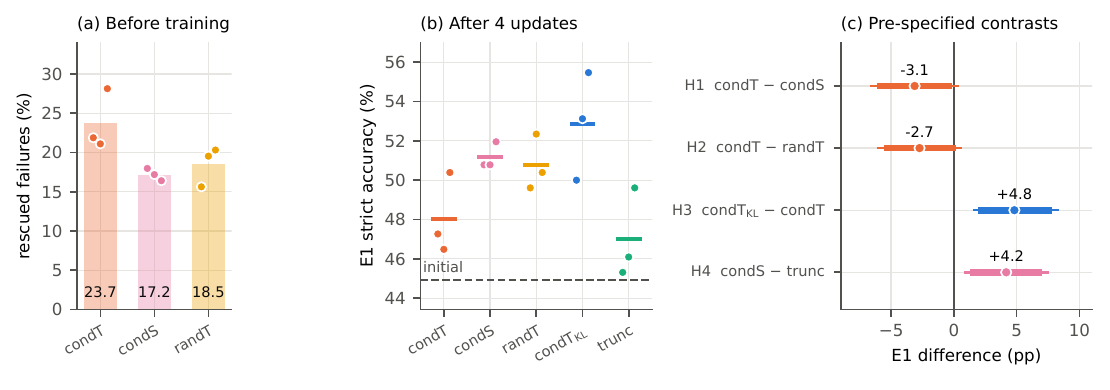}
\caption{Conditional-intervention pilot (exploratory). (a) Failed training trajectories whose final answer is strictly correct before any training; bars are pooled rates, dots are seeds. (b) E1 strict accuracy after 4 updates; dots are seeds, bars are means, and the dashed line is the initial student. (c) Pre-specified contrasts with 90\% (thick) and 95\% (thin) question-level bootstrap intervals.}\label{fig:cond}
\end{figure}

\paragraph{The teacher repairs answers but, as context, trains a worse student.} The manipulation check passed. Before training, condT trajectories ended in a strictly correct answer for \cvRescueT{} of \cvRescueN{} failures (\cvRescueTPct\%), against \cvRescueR{} for randT and \cvRescueS{} for condS (Table~\ref{tab:cond}, Figure~\ref{fig:cond}a). After training the order reversed. condT gained least among the arms with a continuation (\cvDeltaT\,pp over the initial student). Contrary to the prediction, it trailed the student's own rewrite in every seed (H1: \cvHOne\,pp, 95\% CI $[\cvHOneLo,\cvHOneHi]$, 90\% interval $[\cvHOneNinetyLo,\cvHOneNinetyHi]$; per seed \cvHOneSeeds{} questions) and trailed randT (H2: \cvHTwo\,pp, $[\cvHTwoLo,\cvHTwoHi]$; Table~\ref{tab:condcontrast}). The gate failed, and by the pre-written rule no follow-up tests this hypothesis. Most of the gain came from the student's continuation: trunc, which drops it, gained \cvDeltaU\,pp, and condS gained \cvDeltaS\,pp (H4: \cvHFour\,pp, $[\cvHFourLo,\cvHFourHi]$). The continuation also carried the lengthening: trunc hit the cap \cvCapU{} times on average, against \cvInitCap{} for $\theta_0$ and \cvCapS--\cvCapT{} for the arms with a continuation.

\paragraph{Here the loss on teacher tokens appeared to matter.} This result did not replicate in a pre-specified test (Section~\ref{sec:v7}); we report the pilot as it was. Also against the prediction, equivalence of the two treatments of the teacher tokens within $\pm5$\,pp was not shown. Adding reverse KL on the teacher segment gave the best arm (condT$_{\mathrm{KL}}$, \cvDeltaK\,pp over the initial student, $[\cvDeltaKLo,\cvDeltaKHi]$). H3 was \cvHThree\,pp ($[\cvHThreeLo,\cvHThreeHi]$; per seed \cvHThreeSeeds{} questions). Its 90\% interval, $[\cvHThreeNinetyLo,\cvHThreeNinetyHi]$, excludes zero and is not inside the $\pm5$\,pp margin under which the main study established equivalence. It did not clearly beat the student's own rewrite: condT$_{\mathrm{KL}}-$condS was \cvKminusS\,pp ($[\cvKminusSLo,\cvKminusSHi]$; \cvKminusSHard\,pp on the hard subset; descriptive, not pre-specified). The gradient geometry moved in the direction this requires (Section~\ref{sec:mechanism}). At the derail point the teacher segment was \cvTokShareB{} of the tokens and \cvKLShareB{} of the KL mass (main study: \mTokShareB{} and \mKLShareB). Its gradient norm was \cvNormRatioB{} of $\|g_{AC}\|$ (main study \mNormRatioB), and its cosine with $g_{AC}$ was \cvCosBrevAC{} (main study \mCosBrevAC). The median cosine between the AC and all-reverse-KL gradients fell from \mCosUpdAll{} to \cvCosUpd{} over \cvGeoN{} updates, an angle of about \cvAngle{} degrees instead of \cvMainAngle. Every update was still clipped (median pre-clip norm \cvGradNorm). The geometry thus makes a difference possible where the main study ruled it out, but an angle this small does not by itself account for a difference of this size. Three seeds on 256 questions cannot exclude that part of it is chance.

\paragraph{Reading.} Teacher text at the derail point is a better state for the next tokens: more continuations from it end correctly. It is not a better training state for a student that must later answer alone. Continuations that follow teacher text taught less than the student's own rewrite of the same failure, and training the student to write the teacher's text as well removed this deficit. One possible explanation is a mismatch between the states used for training and those reached by the student alone. This is consistent with the concern motivating DAgger~\cite{dagger}, which queries expert labels at learner-induced states; we did not measure state occupancy or test this explanation directly. In the main study the teacher paragraph mattered in neither role. In this pilot it did not help as a state, and the loss on its tokens appeared to matter, but the pilot does not show that a teacher at the derail point beats the student's own rewrite, and the pre-specified replication of Section~\ref{sec:v7} detected no benefit from the loss on teacher tokens. The pilot is exploratory: three seeds, 256 evaluation questions, four contrasts without a multiplicity correction, and no significance claims.
\FloatBarrier

\section{Pre-specified replication at the derail point}
\label{sec:v7}
\paragraph{Design.} H3 was the one pilot result that contradicted the main study, and it was found on the same 256 questions as three other contrasts. We therefore tested it again, as a new hypothesis, in a separately pre-specified study (\texttt{protocol\_v7.txt}; configuration sha256 e4c9d114\ldots, frozen after a smoke test and before its formal trained models were evaluated). It uses \rvPool{} training questions not used anywhere else in this study, \rvNseeds{} seeds and all 512 E1 questions. The initial student failed \rvFail{} of the training questions (\rvFailRate\%). Derail and control points follow the pilot's rule: the derail point started at a median token \rvDerailStart{} (a median \rvDerailRel{} of the way through the answer), with median paragraph KL \rvDerailKL{} against \rvControlKL{} at the control point. Each seed trains on its own block of \rvNtrain{} failures. The teacher samples once from the derail point (cap 832), and the sample is cut offline into nested segments. T64 follows the pilot's rule (up to the first paragraph boundary after 64 tokens, cap 384; median \rvTeachLenSixtyFour{} tokens), and T512 applies the same rule after 512 tokens (cap 832; median \rvTeachLenFiveTwelve{} tokens). All \rvNestN{} pairs of segments were nested. The student then continues each trajectory to EOS or to 4096 tokens in total. Every arm starts from $\theta_0$ and takes 4 AdamW updates of 32 trajectories, as before. The arms differ only in the weight $w$ of the reverse KL on the teacher segment. The student segments always receive reverse KL, and the pre-clip gradient is $g_{\mathrm{stu}}+w\,g_T$ under the same $1/L$ normalization. T64$_{\mathrm{AC}}$ ($w=0$) is the pilot's condT, T64$_{\mathrm{KL}}$ ($w=1$) is its condT$_{\mathrm{KL}}$, and T64$_{\mathrm{W4}}$ ($w=4$) is new. Evaluation is that of the main study, with a fresh common-random-number stream, so that no model shares per-question sampling luck with a pilot model.

\paragraph{Hypotheses.} P1 retests H3: T64$_{\mathrm{KL}}-$T64$_{\mathrm{AC}}>0$. P2 is a weight dose: T64$_{\mathrm{W4}}-$T64$_{\mathrm{KL}}>0$. If the loss on teacher tokens helps because it adds a teacher component to the gradient, a larger weight should help more. The weight changes only that component, not the trajectories. Both are two-sided sign-flip tests on all 512 E1 questions, with Holm correction at 0.05. A result counts as confirmed only if it passes Holm with a positive mean; P1 must also be positive on E1 257--512, the half that no pilot model was evaluated on. Secondary units ran in a pre-specified order under a score-blind time rule: the length dose (T512, then T192), an exact-FP32 re-run of the main study's AC and allRKL, and a fresh condS. P2 was first written as a length dose. We replaced it before any GPU run, because a longer teacher segment also changes what the teacher writes (Section~\ref{sec:deviations}).

\input{tables_v7_auto_preprint}

\begin{figure}[!htbp]\centering
\includegraphics[width=\linewidth]{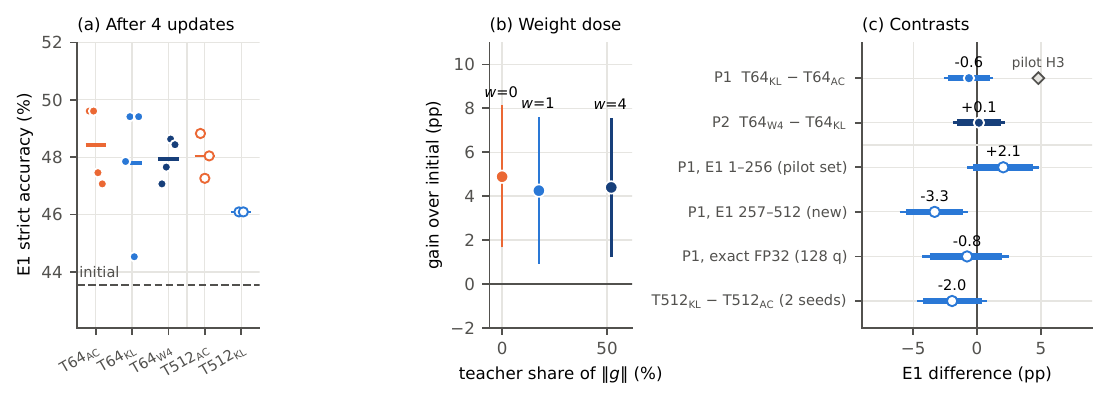}
\caption{Pre-specified derail-point replication. (a) E1 strict accuracy (512 questions); dots are seeds, bars are means, open markers are the secondary length dose, and the dashed line is the initial student. (b) Gain over the initial student against the teacher term's median share of the pre-clip gradient norm, for teacher-segment weights $w=0$, 1 and 4 on the same trajectories (95\% intervals). (c) Contrasts with 90\% (thick) and 95\% (thin) intervals; filled markers are the pre-specified family, and the grey diamond is the pilot's H3.}\label{fig:v7}
\end{figure}

\paragraph{Neither hypothesis was confirmed.} P1 was \rvPOne\,pp (95\% CI $[\rvPOneLo,\rvPOneHi]$, $p=\rvPOneP$; Table~\ref{tab:v7contrast}, Figure~\ref{fig:v7}c). The pilot's $+4.8$\,pp lies outside this interval. P2 was \rvPTwo\,pp ($[\rvPTwoLo,\rvPTwoHi]$, $p=\rvPTwoP$). All three T64 arms gained a similar amount over the initial student: \rvDeltaAC, \rvDeltaKL{} and \rvDeltaW\,pp for $w=0$, 1 and 4, every interval excluding zero (Table~\ref{tab:v7}). The weight changed the gradient as intended. The teacher term's share of the pre-clip gradient norm rose from 0 to \rvShareKL{} and \rvShareW, and the gradient turned \rvAngleKL{} and \rvAngleW{} degrees away from the context-only gradient on the same batch (Figure~\ref{fig:v7}b). Accuracy did not follow. On the pilot's 256 questions P1 was \rvPOneFirst\,pp ($[\rvPOneFirstLo,\rvPOneFirstHi]$), and on the other 256 it was \rvPOneSecond\,pp ($[\rvPOneSecondLo,\rvPOneSecondHi]$). These halves are secondary, and we do not interpret the difference between them. The replication did not support H3; selection noise is a plausible explanation for the largest of four pilot contrasts on 256 questions.

\paragraph{Longer teacher segments (secondary).} Before training, longer teacher segments repaired more failures: \rvRescueS, \rvRescueSixtyFour, \rvRescueOneNineTwo{} and \rvRescueFiveTwelve{} of \rvRescueN{} trajectories ended in a correct answer with no teacher, T64, T192 and T512. A T512 segment contained a final \texttt{\textbackslash boxed} answer in \rvBoxedFiveTwelve{} cases and ended with EOS in \rvTeachEosFiveTwelve. Training on T512 trajectories with the teacher as context gained as much as T64 (\rvDeltaLongAC\,pp, \rvNsLongAC{} seeds). With reverse KL on the long segment, which then carried \rvShareLong{} of the gradient norm and turned the gradient by \rvAngleLong{} degrees, the difference was \rvLongDelta\,pp ($[\rvLongDeltaLo,\rvLongDeltaHi]$). The length interaction was \rvLongInter\,pp. Both rest on \rvLongSeeds{} seeds after one unit was lost (Section~\ref{sec:deviations}), so we read them descriptively: there is no sign that the loss on teacher tokens helps more when the segment is longer.

\begin{figure}[!htbp]\centering
\includegraphics[width=\linewidth]{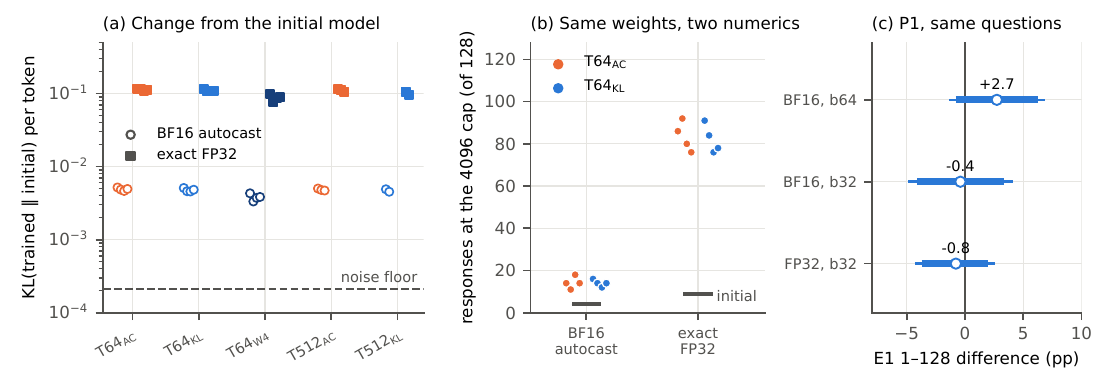}
\caption{BF16 autocast versus exact FP32 for the same trained weights. (a) Per-token KL between each trained model and the initial student on 64 fixed responses (open: BF16 autocast; filled: exact FP32); the dashed line is the KL between the initial student under the two numerics. (b) Responses hitting the 4096-token cap on the first 128 E1 questions, at batch 32 in both numerics; dots are models and the bar is the initial student. (c) P1 on the same 128 questions under three evaluation settings, with 90\% and 95\% intervals.}\label{fig:v7num}
\end{figure}

\paragraph{In exact FP32 the trained models stop terminating.} Unless explicitly labelled exact FP32, evaluations run FP32 weights under BF16 autocast. A 4-step update moves each weight by about $10^{-5}$, below half the BF16 spacing of typical weights, and \rvReverted{} of the changed elements round back to their initial BF16 values. We measured what this does to the model's outputs (Table~\ref{tab:v7num}, Figure~\ref{fig:v7num}a). On 64 fixed responses, the per-token KL between each trained model and the initial student was \rvKLbf{} under BF16 autocast and \rvKLfp{} in exact FP32, \rvKLfold{} times larger; the numerical noise floor is $\rvKLnoise$. Under BF16, the change in log-probabilities had \rvNormRatioPct\% of its FP32 norm, in a similar direction (cosine \rvDeltaCos). The FP32 models also behave differently (Table~\ref{tab:v7fp}, Figure~\ref{fig:v7num}b). We evaluated the \rvSubUnits{} T64$_{\mathrm{AC}}$ and T64$_{\mathrm{KL}}$ models on the first 128 E1 questions in exact FP32 and, at the same batch size, under BF16. Under BF16 they hit the cap \rvSubBbCap{} times and answered \rvSubBbCorrect{} questions correctly (initial student: \rvSubInitBbCap{} and \rvSubInitBbCorrect). In exact FP32 they hit the cap \rvSubFpCap{} times and answered \rvSubFpCorrect{} correctly (initial student: \rvSubInitFpCap{} and \rvSubInitFpCorrect). Every one of the \rvSubUnits{} models did worse in FP32. The initial student did not: on the main study's random stream it hit the cap \rvSubInitOldFpCap{} times in FP32, against \rvSubInitOldBbCap{} under BF16. The truncated FP32 responses are the endless self-corrections of Section~\ref{sec:collapse}. After 4 updates, the trained FP32 weights have therefore already lost the ability to stop. Under BF16 inference most of the update is rounded away, and what remains behaves like a smaller update that helps. Neither numerical setting showed a significant P1 effect (\rvFpPOne\,pp in FP32 and \rvFpPOneBb\,pp under BF16 on the same 128 questions; Figure~\ref{fig:v7num}c), but in FP32 both arms are collapsed models.

\paragraph{Reading.} At the derail point, pre-specified tests detected no benefit from treating teacher tokens as a target (P1) or increasing their weight (P2); the secondary length comparison also showed no benefit. These tests do not establish equivalence. The pilot's H3 does not replicate. The pilot's other finding, that teacher text as context trained a worse student than the student's own rewrite, was not retested: the fresh condS arm came last in the order and did not run. The FP32 result changes how every outcome in this paper should be read. The gains we measured belong to the model as deployed under BF16 inference, which keeps less than a tenth of the update's effect on the output distribution; the full-precision weights after the same 4 updates no longer terminate. Whether arms that are equivalent under BF16 are also equivalent at a dose that is safe in full precision is open.
\section{Termination collapse at a higher dose}
\label{sec:collapse}
\begin{figure}[!htbp]\centering
\includegraphics[width=\linewidth]{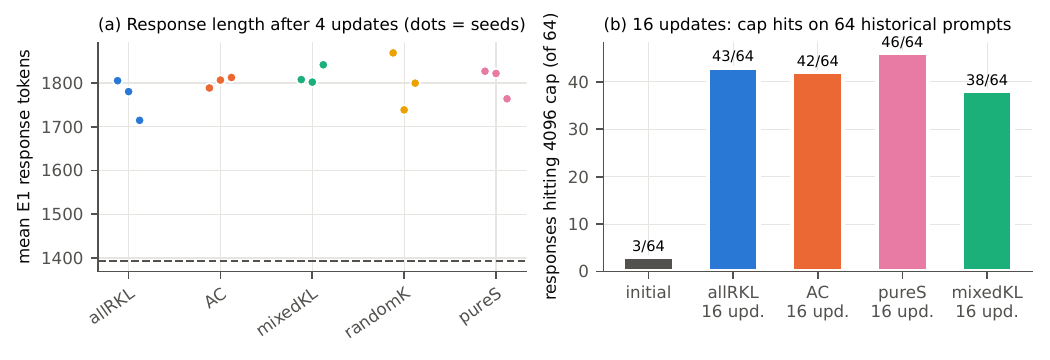}
\caption{(a) Mean E1 response length after 4 updates; the dashed line is the initial student. (b) Responses hitting the 4096-token cap on 64 historical questions under the recommended sampling, after 16 updates.}\label{fig:len}
\end{figure}

In the first, paused version of the study (16 updates of 8 trajectories), seed-1 allRKL hit the 4096-token cap on 42--43 of 64 historical questions under both greedy decoding and the recommended sampling, against 3 for the initial student (Figure~\ref{fig:len}b). With 4 updates of 32 trajectories, the students lengthened their responses but kept terminating. These schedules differ in both optimizer-step count and batch size; gradient clipping does not establish equal AdamW parameter-step lengths.

To ask whether the teacher bridge causes the collapse, we trained AC, mixedKL and pureS on the seed-1 trajectories at the original dose (16 updates of 8) and measured termination on the same 64 historical questions under the same sampling. All collapsed alike: 42 (AC), 38 (mixedKL) and 46 (pureS) responses hit the cap, against 43 for allRKL and 3 for the initial student (Figure~\ref{fig:len}b). Trajectories with no teacher paragraph collapsed at least as much as those with one. The confirmatory scan locates the onset between 4 and 8 updates of 32 (Section~\ref{sec:confirm}), with AC and pureS again collapsing alike. These onsets describe the model under BF16 inference. In exact FP32, the tested T64 replication models trained with 4 updates of 32 already hit the cap on \rvSubFpCap{} of 128 questions (Section~\ref{sec:v7}), so the full-precision weights collapse earlier than the deployed ones. The pure-student control shows that a teacher-written bridge is not necessary for this collapse in our setting.

\section{Online training for 100 steps (exploratory)}
\label{sec:online}
\paragraph{Question and design.} The collapse limited every experiment above to at most 8 updates. We therefore asked two questions. Can a repaired recipe train online for 100 steps without collapsing? And where does the collapse come from: the EOS position, the step size, or the teacher? This part is a recipe search with a pre-specified gate (\texttt{protocol\_v5\_stage01.txt}). It uses a new 256-question dev set drawn from the DeepMath training pool, plus historical questions, and it never touches E1 or E2. Each step samples \wvRows{} new training questions. The current student writes $A$ (cap 256, cut at the first paragraph boundary), the teacher writes one paragraph (cap 384), and the student continues to EOS or a 2048-token total. Each step makes one AC update: reverse KL on $A$ and $C$ and no loss on $B$, with the optimizer and clip of the main study. All recipes see the same question order. Checkpoints are evaluated at steps 25 and 50 on the first 128 dev questions and at step 100 on all 256, with the evaluation of Section~\ref{sec:controlled} (cap 4096). A pre-specified rule aborted a recipe once its dev128 cap hits exceeded three times the initial count (\wvAbortCap). At step 100 the gate required at most \wvGateCap{} cap hits, at least \wvGateCorrect{} correct answers (initial $+3$\,pp), and no decline after step 50.

\begin{figure}[!htbp]\centering
\includegraphics[width=\linewidth]{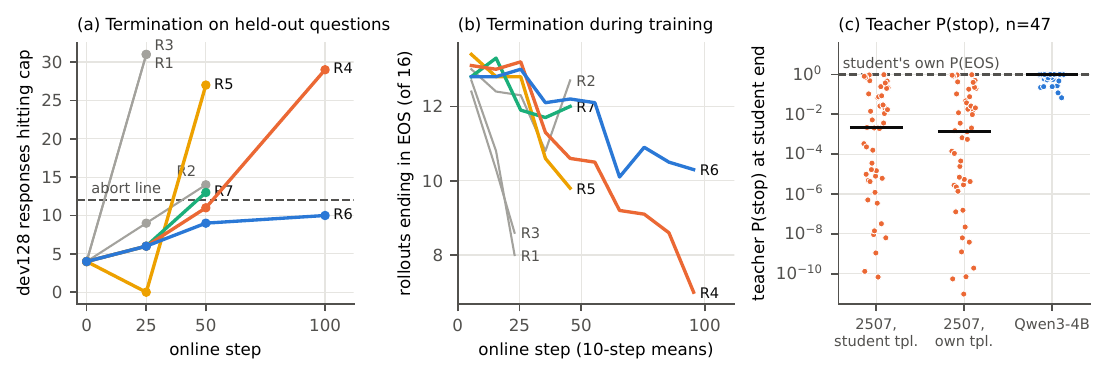}
\caption{Online recipe search (exploratory). (a) Dev responses hitting the 4096-token cap (first 128 questions at every checkpoint); dashed: the pre-specified abort line. (b) Training rollouts that end in EOS, per step of \wvRows. (c) Teacher stop probability at the \wvSzN{} natural end states of the initial student; bars are medians, the dashed line is the student's own median P(EOS). R1--R3 are grey.}\label{fig:online}
\end{figure}

\paragraph{Stopping preferences and final-position masking.} We took the initial student $\theta_0$ and \wvSzN{} natural end states of pure-student trajectories. Here $P(\mathrm{stop})$ sums the probabilities of EOS token IDs 151643 and 151645; the distillation objective continues to treat them as distinct vocabulary items. There, the student's median P(EOS) is \wvSzPS{} and the teacher's is \wvSzPT; the teacher is lower at every state. Reverse KL at the EOS position is \wvSzKLeos, against a mean of \wvSzKLrest{} elsewhere. We then retrained the original collapsing recipe (16 updates of 8). It drives the student's P(EOS) at these states from \wvSzPS{} to \wvSzUnmaskEight{} by update 8 and to \wvSzUnmaskSixteen{} by update 16. Masking only the final EOS position slows this (\wvSzMaskEight{} at update 8) but does not stop it (\wvSzMaskSixteen{} at update 16). The masked student still hit the cap on \wvSzMaskCap{} of \wvSzNHist{} historical questions, against \wvSzUnmaskCap{} unmasked and \wvSzInitCap{} for $\theta_0$. The pre-written rule credited the EOS position only if the masked student hit the cap on 6 or fewer. Removing loss at the final EOS position alone was therefore insufficient; the remaining pre-EOS losses still induced stopping failure. A second check found that the low total stopping probability persisted under the teacher's own chat template. The student's template opens the reply with an empty think block; the teacher's does not. Re-encoding the prompts with the teacher's own template lowers its median P(stop) at the same states to \wvSbTpl. At its own natural end states, the teacher does stop (median \wvSbOwnEnd). The teacher also generates longer answers on these prompts: its own answers average \wvSbOwnLen{} tokens, and \wvSbOwnCap{} of 32 hit the cap (Figure~\ref{fig:online}c).

\paragraph{No repaired recipe passed.} Table~\ref{tab:v5} lists all seven recipes. At learning rate $10^{-6}$, the student collapsed by step 25 with or without EOS masking. At $3\times10^{-7}$, the masked recipe aborted at step 50. The unmasked one (R4) completed 100 steps and gained accuracy (\wvRfourCorrect{} vs.\ \wvInitCorrect{} correct). It still failed the gate with \wvRfourCap{} cap hits, against \wvInitCap{} for $\theta_0$. Its training rollouts that ended in EOS fell from \wvRfourEosFirst{} to \wvRfourEosLast{} of \wvRows{} per step (Figure~\ref{fig:online}b). EOS anchoring (R5) adds $-\log p_\theta(\mathrm{EOS})$ at the final state of every rollout that ends in EOS. Over the last 25 updates it kept the median of that probability above \wvRfivePEos{} (per-update minimum \wvRfivePEosMin), and it briefly removed every truncation (\wvRfiveCapTwentyFive{} cap hits at step 25). By step 50 there were \wvRfiveCapFifty. Keeping the student willing to stop at the states where it ends does not keep it reaching those states. Reference-KL regularization toward $\theta_0$ (R7, using the same numerical coefficient 0.5 reported for some SimpleOPD settings~\cite{simpleopd}) had no visible effect: \wvRsevenCapFifty{} cap hits at step 50, against \wvRfourCapFifty{} for R4. Its scaled gradient was \wvRsevenRatioFirst\% of the distillation gradient's norm at step 1 and \wvRsevenRatioLast\% at step 50. The median cosine between the two was $\wvRsevenCos$ from step 10 on: the two gradients opposed one another, and reference regularization at this coefficient did not prevent collapse. This is not a replication of SimpleOPD's sampled-token PPO recipe. Coefficients do not transfer across objectives: our distillation term is a full-vocabulary KL, far larger per token than the early reference term. We did not test stronger regularization.

\paragraph{A length-matched teacher.} Qwen3-4B (the hybrid model, in non-thinking mode) shares the student's chat template and vocabulary. At the same \wvSzN{} end states its median P(stop) is \wvScMed{}, with \wvScFrac\% of states at least 0.5. Its own answers average \wvScOwnLen{} tokens, and \wvScOwnCap{} of 32 hit the cap. We wrote two readings in advance. If this teacher's P(stop) is high and R6 passes the no-collapse criterion while R4 does not, the length mismatch is supported. If P(stop) is high but R6 still collapses, the mismatch is not the only cause. R6 completed 100 steps with \wvRsixCap{} cap hits (median length \wvRsixMedLen{} tokens, against \wvRfourMedLen{} for R4 and \wvInitMedLen{} for $\theta_0$). That exceeds the limit of \wvGateCap, so the swap did not meet the pre-written no-collapse criterion. Descriptively, the swap removes most excess truncations, but it changes teacher strength together with stopping preference. The comparison is paired by question and evaluation random stream. \wvCapFourOnly{} questions hit the cap only under R4 and \wvCapSixOnly{} only under R6 (exact sign test $p<\wvCapP$). Truncations in excess of $\theta_0$'s fell from \wvExcessFour{} to \wvExcessSix, a \wvExcessRed\% reduction.

\paragraph{The gain left with the collapse.} R6 also lost R4's accuracy gain: \wvRsixCorrect{} correct against \wvRfourCorrect{} for R4 and \wvInitCorrect{} for $\theta_0$ (discordant \wvCorFourOnly{} vs.\ \wvCorSixOnly, $p=\wvCorP$). On the \wvSubN{} questions that no model truncated, $\theta_0$, R4 and R6 answered \wvSubInit, \wvSubFour{} and \wvSubSix{} correctly. This subset is selected on outcomes after training, so it is descriptive only. In this pair, switching away from the verbose teacher reduced both the accuracy gain and truncation. The swap changes the teacher's strength as well as its length preference, because Qwen3-4B in non-thinking mode is a weaker mathematical teacher than 2507. R6 therefore cannot tell whether a teacher as strong as 2507 but matched in length would give the gain without the collapse. Each recipe ran once.

\input{tables_v5_auto_preprint}

\section{Related work}
\paragraph{Teacher-written segments in OPD.} TCOD-B2F~\cite{tcod} conditions on a prefix of a successful teacher trajectory, trains only the student's continuation, and shrinks the prefix over a curriculum. ReOPD~\cite{reopd} replays teacher histories and distills the current student action. These share AC's context-only role for teacher text, but their rollout procedures and loss estimators differ from ours. Guided-OPD~\cite{guided} randomly assigns turn roles under a decaying teacher curriculum, using reverse KL on student turns and forward KL on teacher turns. This shares mixedKL's role-specific divergence directions, not its exact objective; Guided-OPD also compares alternative KL assignments in Table~2(a). DASH-OPD~\cite{dash} switches executors using discrepancy and hysteresis, with losses on valid generated response tokens from both models. Relay-OPD~\cite{relay} uses the same model pair on mathematical reasoning and compares three objectives at teacher-leg positions, including a saved-student-draft alternative. The student may resume after a teacher leg; the rollout ends after the final budgeted leg. FutureBridge-OPD~\cite{futurebridge} selects a turn with a large mean sampled $\log(p_{\mathrm{student}}/p_{\mathrm{teacher}})$ and accepts a bridge when the bridge-induced student continuation has a higher teacher-preferred-token ratio over a bounded horizon than the paired student continuation. Accepted bridge tokens are supervised, and its random-turn ablation performs worse. The gate is a teacher-based proxy, not a task-reward test. We compare loss treatments on identical trajectories and relate the outcome to author-separated gradient geometry.
\paragraph{Triggered intervention.} MAESTRO~\cite{maestro} uses the same model pair on mathematical reasoning and adapts intervention timing and depth to paragraph-level disagreement. Relay-OPD's delayed-handover probe and trigger-stop training ablation test placement and truncation; adding a teacher leg outperforms truncating at the trigger under the matched one-handover budget. Our trunc arm is a related control, under a different protocol. Seg-OPD~\cite{segopd} selects high-entropy segments from failed responses and trains a preference for teacher redrafts over original segments, with position-matched random anchors in both a probe and a training ablation. UOPD~\cite{uopd} triggers on low teacher support for student actions or a student--teacher log-probability gap. Its online ablations remove teacher-action supervision ($\beta=0$), remove teacher execution, vary the supervision weight, and randomize selection while matching the target intervention-rate schedule. These precedents already distinguish several roles of intervention. Our pilot instead holds the realized teacher text fixed for the loss comparison and measures author-separated gradients; its apparent benefit from teacher-token loss did not replicate in the pre-specified test.
\paragraph{Gradient analyses.} Jiang et al.~\cite{sparsedense} analyze reward--distillation interference with cross-signal neural tangent kernels and a first-order loss decomposition at student-generated positions; the split is by training signal rather than token author. Prefix-RFT~\cite{prefixrft}, Appendix~A.3, reports larger total gradient norms when a small fraction of demonstration-prefix tokens are updated rather than frozen; this is not a direct measurement of their author-separated norm share. Seg-OPD~\cite{segopd}, Appendix~B.3, already measures pre-clipping parameter-gradient norms of token, preference and redraft-SFT losses, and FutureBridge-OPD~\cite{futurebridge} gives formal bridge-gradient results. Other token-level analyses study sampled-gradient estimation noise~\cite{onepercent} and directional token mismatch~\cite{tide}. PCGrad~\cite{pcgrad} projects conflicting parameter gradients; LMOPD~\cite{lmopd} instead applies a one-sided, priority-aware projection to centered log-policy corrections. These conflict-handling ideas motivated our pre-specified projection rule. Our pilot found no negative segment-gradient cosine meeting that rule, so the projection arm was excluded.
\paragraph{Length inflation and collapse in OPD.} Yang et al.~\cite{eosdisagree} identify termination-token mismatch as an important source of inflation in base-to-post-trained distillation, while also observing residual inflation after correction. Our probe sums both EOS probabilities, but our full-vocabulary loss does not merge their semantics; we did not run that objective ablation and do not exclude this mechanism. SimpleOPD~\cite{simpleopd} reports that structural termination-token masking mitigates but does not remove length growth, whereas reference-KL regularization nearly removes truncation in its Intern-S2 setting. Our final-EOS-position mask was also insufficient; reference regularization at the same numerical coefficient had no visible effect under our different full-vocabulary objective. Stable-OPD~\cite{stableopd} combines student-to-reference KL with off-policy golden-solution SFT alongside student rollouts. Fu et al.~\cite{revisitopd} present qualitative cases where the teacher favours continuation after an answer is available; our probe measures a related stopping disagreement in our model pair. Wang et al.~\cite{demystopd} describe long- and short-response exploitation under sampled-token OPD, and Cui et al.~\cite{gainscollapse} interpret the collapse in their runs as reward hacking of the teacher's implicit reward. We add a final-EOS-position masking test, a template check, and a teacher swap that examines stopping preference while also changing teacher strength.
\paragraph{Foundations.} GKD~\cite{gkd} trains on student-generated sequences and permits alternative token-distribution divergences. MiniLLM~\cite{minillm} uses reverse KL with per-token teacher--student mixture sampling. DAgger~\cite{dagger} aggregates expert-labelled states visited by a mixture of learner and expert policies.

\section{Deviations from the initial protocol}
\label{sec:deviations}
Deviations (i)--(iv) were made before any arm-versus-arm comparison was computed, and each is logged with its timestamp.
(i) The evaluation decoding changed from greedy to the sampling Qwen3 recommends, after trained students failed to terminate under greedy decoding. A pre-written decision test did \emph{not} favour sampling: it did not halve cap hits. The change follows the model card and the authors' decision.
(ii) The dose was reduced from 16 updates of 8 trajectories to 4 updates of 32 (Section~\ref{sec:collapse}).
(iii) The primary E1 set was reduced from 512 to the first 256 questions to fit the compute budget.
(iv) We ordered the fixed-sequence family poorly: its first contrast was the one our own mechanism predicted to be null. The family therefore stopped at step one, and the AC vs.\ pureS contrast could only be reported descriptively. We addressed this with the separately pre-specified test of Section~\ref{sec:confirm}, where that contrast was primary.
(v) In the pilot of Section~\ref{sec:cond}, building the seed-2 trajectories crashed inside the frozen grader, which could not print a predicted integer with more than 4{,}300 digits (a Python safety limit). Sampling is seeded, so the retry would have crashed at the same place. We stopped it and relaunched the driver with the limit disabled (\texttt{PYTHONINTMAXSTRDIGITS=0}), before any seed-2 outcome existed but after the seed-1 arms had been evaluated. It is a technical fix, not a design change: no frozen file changed, the grading rules are unchanged, and the units already completed never reached this code path.
(vi) In the replication of Section~\ref{sec:v7}, P2 was first written as a length dose, $[\Delta(512)-\Delta(64)]$ with $\Delta(d)$ the T$d_{\mathrm{KL}}-$T$d_{\mathrm{AC}}$ difference. On 2026-10-20, before any of its GPU runs, we replaced it by the weight dose, because the length dose also changes the teacher's content and a difference of differences has less power; the length dose became secondary. The same day, also before the configuration was frozen, we corrected one protocol sentence that quoted a hardware-dependent value of the probe cosine from a CPU unit test. (vii) The original machine was unavailable, so the replication ran on another instance with the same GPU type, the same software versions and the same model revisions, which we verified file by file against the official manifests. (viii) One secondary unit, the seed-3 T512$_{\mathrm{KL}}$ model, was killed by the operating system during the grading of its E1 responses, most likely because one response exhausted memory in the grader; the cause could not be checked before the machine was shut down. The score-blind time rule then skipped its retry and every later unit (seed-4 T512, the FP32 re-run of the main study, T192 and condS), as the protocol specifies. The pre-specified tests were complete before this happened.

\section{Relation to prior results}
Our equivalence result concerns three loss treatments on identical fixed trajectories; it does not establish equivalence of the complete methods that motivated them. TCOD-B2F and ReOPD use teacher text as context, whereas Guided-OPD and Relay-OPD also compare supervision choices within their own online procedures. Guided-OPD's KL-direction ablation and UOPD's teacher-loss ablation report benefits from their default supervision choices. Our null result is conditional on a short, fixed-position bridge, dose and evaluation precision, so it cannot overturn those results. When the teacher writes at the student's derail point, our pilot suggested a benefit from teacher-token loss (Section~\ref{sec:cond}), but the pre-specified replication detected neither this difference nor a benefit from fourfold weighting (Section~\ref{sec:v7}). This tests the supervision role shared with several prior methods rather than their full recipes.

Our confirmatory test also detected no state benefit from a single-paragraph bridge (Section~\ref{sec:confirm}). Five differences limit comparison. (i) \emph{Dose}: our main comparison takes 4 updates; the cited methods generally use longer training schedules, whose state benefits may accumulate. (ii) \emph{Bridge size and placement}: our bridge is one paragraph, 2.7\% of the tokens, inserted after a student prefix, whereas TCOD-B2F uses prefixes of successful teacher trajectories and Relay-OPD places corrective legs at triggers. Our longer derail-point segment (median \cvTeachLen{} tokens, \cvTokShareB{} of tokens) also showed no state benefit (Section~\ref{sec:cond}); increasing the median segment length to \rvTeachLenFiveTwelve{} tokens (\rvGeoTokLong{} of trajectory tokens) gave no significant accuracy benefit in the replication (Section~\ref{sec:v7}). (iii) \emph{Task}: TCOD, Guided-OPD, DASH-OPD, FutureBridge-OPD and UOPD study multi-turn agents; Relay-OPD and MAESTRO also study mathematical reasoning with our model pair. Task domain alone therefore cannot explain the difference. (iv) \emph{Dose ceiling}: with fixed trajectories, termination collapses before a larger dose can be evaluated usefully. With fresh rollouts and a 10$\times$ smaller learning rate, the student reached 100 steps and gained accuracy but still stopped terminating. A teacher swap reduced most excess truncations and removed the gain while also changing teacher strength (Section~\ref{sec:online}). (v) \emph{Numerics}: our primary results use BF16 inference, which rounds away most of a 4-update change; the tested replication weights already collapse in exact FP32 (Section~\ref{sec:v7}). We did not compare effective evaluation precision across papers. With our one-paragraph bridge and at most 8 clipped updates under BF16 inference, we detected no accuracy benefit from the bridge; at the derail point, we likewise detected no benefit from teacher-token loss, its weight or the segment length. Only the main study's three loss treatments were tested for equivalence within a pre-specified margin. How bridge fraction and a training regime that preserves termination affect these roles remains open in our setting.

\section{Verification}
\label{sec:verify}
All outcomes were re-derived independently from the raw non-weight archive by a separate checker that did not run any archived code. It regraded 11{,}520 responses with the frozen graders (0 mismatches), recounted every model, reimplemented the paired bootstrap (identical means; intervals within 0.26\,pp; the TOST conclusions unchanged) and recomputed the mechanism medians from the update logs (identical). One grader call on a question outside the analysed set timed out non-deterministically, which affects no reported number. The confirmatory experiment was verified the same way. All 13{,}824 of its responses regraded with 0 mismatches. Every checkpoint used exactly the first 512 frozen E1 questions. The training schedule matched the protocol: each pass covered every trajectory once, and AC and pureS used the same order at every step. Recomputed means were identical and interval endpoints were within 0.13\,pp. The check found one reporting error: the frozen script printed a dose-2 Holm $p$ of 0.83 without the monotonicity step (correct value 1.00). It is corrected in Table~\ref{tab:v4contrast} and changes no decision. The online follow-up (Section~\ref{sec:online}) was checked the same way from its archive. All 2{,}368 of its responses regraded with 0 mismatches. Every count in Table~\ref{tab:v5}, all derived values in Section~\ref{sec:online} and every abort and gate decision were reproduced. The check found one grader false negative: the frozen grader times out on one gold answer ($2^{2009^2}$), so 6 mathematically equal answers are graded wrong. Crediting them would raise R4 at step 100 from \wvRfourCorrect{} to 133, and the R4--R6 correctness comparison would become 30 vs.\ 16 discordant ($p=0.054$). No abort or gate decision changes, and we report the frozen grader's verdicts as the protocol requires. The conditional pilot (Section~\ref{sec:cond}) was checked the same way from its archive. A later regrade with the frozen grader covered all 5{,}866 of its responses (E1 evaluations, failures and built trajectories) with 0 mismatches; nine grader calls timed out and were stored as incorrect, as the frozen grader specifies, and none of them becomes correct with a 60-second timeout. Every count, every bootstrap mean and win/loss count, the gradient-geometry medians and the response-for-response match of the initial evaluation were reproduced; interval endpoints agreed within 0.3\,pp. The check found no numerical error. It flagged wording that went beyond the contrasts, for example describing the teacher as helpful as a target when the pilot does not show that it beats the student's own rewrite; we corrected it. The replication of Section~\ref{sec:v7} was checked the same way from its archive, by a checker that re-implemented the statistics. It regraded all 12{,}352 stored E1 responses (including 448 from the unit that was later lost) and all 2{,}048 built trajectories with the frozen grader, with 0 mismatches in any stored score field, and every integrity check passed (question order, batch-to-score concatenation, decoding, stop reasons and lengths). Every count, rescue count, gradient-geometry median, probe summary and time-rule decision was reproduced. Every contrast mean was identical, interval endpoints agreed within 0.4\,pp, and both pre-specified decisions matched. Two notes: one stored correct answer takes 1.1--1.4 seconds against the grader's 2-second limit and was graded wrong once under parallel load, which would move P2 by 0.05\,pp; and an older version of the parser's dependency changes the parsed answer, but not the grade, of three responses.

\section{Limitations}
\begin{itemize}
\item \textbf{Scale.} One student--teacher pair, 128 training questions, at most 8 updates of 32 per arm. Larger doses change the student qualitatively (Section~\ref{sec:collapse}). An exploratory 100-step online search found no recipe that avoids the collapse with this teacher (Section~\ref{sec:online}); other model pairs are untested.
\item \textbf{Online follow-up.} Each recipe ran once, the dev set is drawn from the training-pool distribution, and the outcomes are exploratory. Its teacher swap changes teacher strength together with length preference. Stronger reference regularization, masking of the last $k$ positions, an objective that merges EOS semantics, and a length-matched teacher as strong as 2507 are untested.
\item \textbf{Conditional pilot.} Three seeds, E1 only (256 questions), four contrasts without multiplicity correction, and exploratory by protocol. The derail point is a single heuristic (the paragraph with the highest mean KL); other locators, larger teacher segments, forward KL on the teacher tokens and longer training are untested. Its H3 result did not replicate (Section~\ref{sec:v7}).
\item \textbf{Seeds.} Question-level bootstrap intervals are conditional on three training runs per arm; they do not estimate the distribution over training seeds.
\item \textbf{Evaluation numerics.} The main evaluation samples once per question with common random numbers, so per-question outcomes are noisy; we rely on three seeds per arm and a matched random perturbation (noiseAC) as a reference. In the decoding pilot before the main study, even greedy BF16-autocast decoding was not batch-invariant (0--3\% identical responses across batch sizes 8, 64 and 128), so greedy decoding would not have removed this noise.
\item \textbf{Effective update.} With FP32 master weights and BF16 matmuls, most of a 4-step $3\times10^{-6}$ update is rounded away inside the matmuls: in all 18 trained models of the main study, 93.5--93.7\% of changed elements round back to their initial BF16 value, and the effective update has cosine 0.30 with the FP32 update. In the replication, BF16 inference kept \rvNormRatioPct\% of the update's effect on log-probabilities, and the full-precision weights hit the length cap on \rvSubFpCap{} of 128 questions (Section~\ref{sec:v7}). The primary accuracy findings therefore concern the model as deployed under BF16 inference; the explicit FP32 check evaluates the same trained weights in full precision. The FP32 evaluation covers only the two T64 arms of the replication, 128 questions and one dose.
\item \textbf{Replication.} One model pair and one dose. The T192 dose, a fresh condS arm and the FP32 re-run of the main study did not run within the time budget, and the length dose has two complete seeds.
\item \textbf{Fixed trajectories.} The main comparison is fixed-trajectory distillation, not on-policy training. The S4 check is a single-seed on-policy replication.
\item \textbf{Evaluation sets.} Public benchmarks may appear in pre-training data. E1 is a held-out DeepMath subset filtered by answer type before any evaluation; E2 is a fixed GSM8K subset, not the full test set.
\end{itemize}

\section{Conclusion}
A teacher segment in an on-policy trajectory can act as a state for the student's next tokens or as a target for the student's own distribution. Measured by author-segmented gradients, a one-paragraph bridge after the student's first paragraph contributes a small component nearly orthogonal to the student-segment gradient, and pre-specified tests detected no clear benefit from using it as context. Placed at the point where failed answers derail, teacher text repaired more answers, but as context it trained a worse student than the student's own rewrite in an exploratory pilot. The pilot's other signal, that the loss on teacher tokens helps there, did not replicate in the pre-specified follow-up, and a fourfold weight on the teacher term changed the gradient direction by \rvAngleW{} degrees without a significant accuracy benefit. The pre-specified tests detected no added benefit from teacher-induced states or from the loss on teacher tokens under the tested dose and BF16 inference. The main study established equivalence of three loss treatments within $\pm5$\,pp; the other null tests do not establish equivalence. Author-segmented gradient measurements are cheap to take, and they predicted the equivalence in the main study, but a larger teacher share of the gradient did not yield a detectable accuracy benefit. Two limits bound all of this. In the fixed-trajectory dose sweeps, gains peaked at 4 updates before termination collapsed; the online teacher swap implicates the teacher's stopping preference but also changes its strength, and under the BF16 inference we used, those 4 updates are largely rounded away: the tested replication weights had already stopped terminating reliably in full precision. Training at a dose that is safe in full precision, and comparing teacher-token treatments there, is the open problem for this pair.

\section*{Acknowledgments}
The GPU servers used in this work were rented by the authors.
AI coding and writing assistants were used for experiment scripting,
analysis tooling, drafting and checking the manuscript. The authors
reviewed all outputs and take full responsibility for the content.

\FloatBarrier
\appendix
\section{Frozen protocol in brief}
\label{app:protocol}
The original main-study plan was recorded in \texttt{protocol\_v2.txt} (design, data hashes, statistics), with the projection-arm decision rule in \texttt{pilot\_rule\_v1.txt}. The outcome-informed amendments in \texttt{protocol\_v3\_amendment.txt} changed decoding, dose and E1 size after observing nontermination in an initial trained arm, but before between-arm accuracy comparisons. These are internal specification records, with access described in Appendix~\ref{app:repro}. The confirmatory family is a fixed sequence on E1 at two-sided $\alpha=0.05$: AC$-$allRKL, then AC$-$mixedKL, AC$-$randomK, AC$-$pureS, each tested only if all earlier contrasts were significant. Equivalence uses TOST at $\pm5$\,pp, i.e.\ the 90\% question-level bootstrap interval must lie inside the margin. Per-question scores are the mean over the three seeds; the bootstrap resamples questions (10{,}000 draws, seed 20261006). E2, the noise control, the on-policy check and the collapse sub-study are descriptive. The main-study amendments used termination and update statistics; they preceded between-arm accuracy comparisons. Table~\ref{tab:config} lists the frozen configuration. The confirmatory experiment of Section~\ref{sec:confirm} was frozen separately in \texttt{protocol\_v4.txt} (configuration \texttt{config\_s3\_v4.json}, sha256 d93ae112\ldots) before any of its checkpoints was evaluated. Its single primary test was AC$-$pureS on E1 at 4 updates. Doses 2 and 8 were secondary with Holm correction, and the dose slope, E2 and termination were descriptive. Its analysis script was hash-frozen in that configuration and verified unchanged before it was run. The online follow-up of Section~\ref{sec:online} was frozen in \texttt{protocol\_v5\_stage01.txt} (configuration sha256 804c7487\ldots) before any of its GPU runs, with the stage-0 rule, the abort rule and the gate. Three amendments followed, each written and hashed into the run configuration before the runs it governs (their header times are approximate; the archived configurations identify the amendments incorporated into each run): R5 (EOS anchoring) after stage 0 and before any stage-1 result; the template check, run in parallel with R2; and R6/R7, written after R1--R5 had failed the gate and therefore exploratory by construction. The teacher weights for R6 are pinned by their file hashes. The conditional pilot of Section~\ref{sec:cond} was frozen in \texttt{protocol\_cond\_pilot.txt} (configuration file sha256 d291d9cb\ldots; its content digest, 1c0df7dd\ldots, is recorded in every completion file) before its formal GPU runs, after a smoke test on 48 questions that changed only the teacher-segment rule (a minimum of 64 tokens, because single paragraphs were often a heading or a rule line). Its analysis script, \texttt{analyze\_cond.py}, was committed before any trained pilot model was evaluated; Section~\ref{sec:deviations}(v) records its one technical amendment. The replication of Section~\ref{sec:v7} was frozen in \texttt{protocol\_v7.txt} (configuration \texttt{config\_v7.json}, sha256 e4c9d114\ldots) after a smoke test and before its formal trained models were evaluated. Its pre-specified family was P1 and P2 with Holm correction at 0.05, plus the requirement that P1 be positive on E1 257--512; the analysis script \texttt{analyze\_v7.py} and every code file were hash-frozen in the configuration. Units ran in a fixed priority order (failures, localization, four builds, the initial evaluation, T64$_{\mathrm{AC}}$/T64$_{\mathrm{KL}}$ with FP32 evaluation, T64$_{\mathrm{W4}}$, T512, the FP32 re-run of the main study, T192, condS), and a unit started only if 1.25 times the median duration of completed units of its kind fit before the deadline. Section~\ref{sec:deviations}(vi)--(viii) records its amendments.

\FloatBarrier
\section{Per-unit results and mechanism medians}
Table~\ref{tab:mech} gives mechanism medians by unit group, Table~\ref{tab:units} every fixed-trajectory unit, Table~\ref{tab:online} the on-policy units and Table~\ref{tab:collapse} the collapse sub-study.
\input{appendix_auto_preprint}
\clearpage
\section{Reproducibility}
\label{app:repro}
The protocols, amendments, run configurations, analysis scripts and complete SHA-256 manifests are retained by the authors and are available on request via the contact emails on the first page. The protocol and configuration filenames cited in this paper identify these internal records. The protocols were not deposited in a public registration service, and this version provides no public artifact download. Hashes identify file contents; they do not independently establish when the plans were written. The experiment-specific timing and outcome-informed amendments are disclosed in Appendix~\ref{app:protocol} and Section~\ref{sec:deviations}.

The non-weight archive contains trajectories, update logs with per-segment gradient statistics, every evaluation response and score, configurations with input hashes, driver event logs and pause records. It is retained locally with SHA-256 digests and may also be requested from the authors. The analysis, tables and figures are generated by scripts from the archived JSON. The independent verification of Section~\ref{sec:verify} was run from this archive.
\end{document}

%% file: macros_auto.tex
\newcommand{\mTokShareB}{2.7\%}\newcommand{\mTokShareBLo}{2.2\%}\newcommand{\mTokShareBHi}{4.1\%}
\newcommand{\mKLShareB}{1.8\%}
\newcommand{\mNormRatioB}{0.080}\newcommand{\mNormRatioBLo}{0.046}\newcommand{\mNormRatioBHi}{0.486}
\newcommand{\mCosBrevAC}{0.126}\newcommand{\mCosBrevACLo}{-0.008}\newcommand{\mCosBrevACHi}{0.247}
\newcommand{\mCosBfwdAC}{0.090}
\newcommand{\mCosUpdAll}{0.997}\newcommand{\mCosUpdAllLo}{0.901}
\newcommand{\mCosUpdMixed}{0.999}\newcommand{\mCosUpdMixedLo}{0.996}
\newcommand{\mGradNorm}{5.333}

\newcommand{\mNupd}{48}

%% file: macros_v4_auto.tex
\newcommand{\vPrimMean}{+0.9}
\newcommand{\vPrimLo}{-1.5}
\newcommand{\vPrimHi}{+3.2}
\newcommand{\vPrimP}{0.48}

\newcommand{\vACfourInit}{+4.5}
\newcommand{\vPSfourInit}{+3.6}
\newcommand{\vCapTwo}{24}
\newcommand{\vCapFour}{41}
\newcommand{\vCapEight}{184}
\newcommand{\vSlope}{+0.26}

%% file: macros_v5_auto.tex
\newcommand{\wvInitCorrect}{120}
\newcommand{\wvInitCap}{8}
\newcommand{\wvInitMedLen}{1108}

\newcommand{\wvGateCap}{16}
\newcommand{\wvGateCorrect}{128}
\newcommand{\wvAbortCap}{12}
\newcommand{\wvSzN}{47}
\newcommand{\wvSzPS}{0.991}
\newcommand{\wvSzPT}{0.002}
\newcommand{\wvSzKLeos}{5.6}
\newcommand{\wvSzKLrest}{0.44}

\newcommand{\wvSzUnmaskEight}{0.32}
\newcommand{\wvSzUnmaskSixteen}{0.000}
\newcommand{\wvSzMaskEight}{0.74}
\newcommand{\wvSzMaskSixteen}{0.001}
\newcommand{\wvSzMaskCap}{34}
\newcommand{\wvSzUnmaskCap}{43}
\newcommand{\wvSzInitCap}{3}
\newcommand{\wvSzNHist}{64}
\newcommand{\wvSbTpl}{0.0013}
\newcommand{\wvSbOwnEnd}{0.993}
\newcommand{\wvSbOwnLen}{2440}
\newcommand{\wvSbOwnCap}{10}
\newcommand{\wvScMed}{0.967}
\newcommand{\wvScFrac}{85}
\newcommand{\wvScOwnLen}{1509}
\newcommand{\wvScOwnCap}{2}
\newcommand{\wvRfourCorrect}{132}
\newcommand{\wvRfourCap}{59}
\newcommand{\wvRfourMedLen}{1932}
\newcommand{\wvRsixCorrect}{119}
\newcommand{\wvRsixCap}{21}
\newcommand{\wvRsixMedLen}{1266}
\newcommand{\wvRfourEosFirst}{13.1}
\newcommand{\wvRfourEosLast}{7.0}

\newcommand{\wvRows}{16}
\newcommand{\wvCapFourOnly}{45}
\newcommand{\wvCapSixOnly}{7}
\newcommand{\wvCapP}{10^{-7}}
\newcommand{\wvExcessFour}{51}
\newcommand{\wvExcessSix}{13}
\newcommand{\wvExcessRed}{75}
\newcommand{\wvCorFourOnly}{30}
\newcommand{\wvCorSixOnly}{17}
\newcommand{\wvCorP}{0.079}
\newcommand{\wvSubN}{187}
\newcommand{\wvSubInit}{108}
\newcommand{\wvSubFour}{127}
\newcommand{\wvSubSix}{109}
\newcommand{\wvRfiveCapTwentyFive}{0}

\newcommand{\wvRfiveCapFifty}{27}
\newcommand{\wvRfivePEos}{0.999}
\newcommand{\wvRfivePEosMin}{0.86}
\newcommand{\wvRsevenCapFifty}{13}
\newcommand{\wvRfourCapFifty}{11}
\newcommand{\wvRsevenRatioFirst}{1.4}
\newcommand{\wvRsevenRatioLast}{13}
\newcommand{\wvRsevenCos}{-0.61}

%% file: macros_cond_auto.tex
\newcommand{\cvGeoN}{24}
\newcommand{\cvTokShareB}{5.2\%}
\newcommand{\cvKLShareB}{9.3\%}
\newcommand{\cvNormRatioB}{0.19}

\newcommand{\cvCosBrevAC}{0.33}

\newcommand{\cvCosUpd}{0.986}

\newcommand{\cvGradNorm}{5.2}
\newcommand{\cvAngle}{10}
\newcommand{\cvMainAngle}{4}

\newcommand{\cvPool}{1{,}067}
\newcommand{\cvFail}{618}
\newcommand{\cvFailNoEos}{25}
\newcommand{\cvFailRate}{58}
\newcommand{\cvNtrain}{128}
\newcommand{\cvNtrainAll}{384}
\newcommand{\cvDerailStart}{500}
\newcommand{\cvDerailRel}{0.41}
\newcommand{\cvDerailParaOne}{2}

\newcommand{\cvDerailKL}{1.47}
\newcommand{\cvControlKL}{0.33}
\newcommand{\cvTeachLen}{81}

\newcommand{\cvRescueT}{91}
\newcommand{\cvRescueTPct}{23.7}

\newcommand{\cvRescueS}{66}

\newcommand{\cvRescueR}{71}

\newcommand{\cvRescueN}{384}

\newcommand{\cvInitCap}{11}

\newcommand{\cvDeltaT}{\ensuremath{+3.1}}

\newcommand{\cvCapT}{33}

\newcommand{\cvDeltaS}{\ensuremath{+6.2}}

\newcommand{\cvCapS}{25}

\newcommand{\cvDeltaK}{\ensuremath{+7.9}}
\newcommand{\cvDeltaKLo}{\ensuremath{+3.8}}
\newcommand{\cvDeltaKHi}{\ensuremath{+12.4}}

\newcommand{\cvDeltaU}{\ensuremath{+2.1}}

\newcommand{\cvCapU}{13}

\newcommand{\cvHOne}{\ensuremath{-3.1}}
\newcommand{\cvHOneLo}{\ensuremath{-6.6}}
\newcommand{\cvHOneHi}{\ensuremath{+0.4}}

\newcommand{\cvHTwo}{\ensuremath{-2.7}}
\newcommand{\cvHTwoLo}{\ensuremath{-6.1}}
\newcommand{\cvHTwoHi}{\ensuremath{+0.7}}

\newcommand{\cvHThree}{\ensuremath{+4.8}}
\newcommand{\cvHThreeLo}{\ensuremath{+1.6}}
\newcommand{\cvHThreeHi}{\ensuremath{+8.3}}

\newcommand{\cvHFour}{\ensuremath{+4.2}}
\newcommand{\cvHFourLo}{\ensuremath{+0.8}}
\newcommand{\cvHFourHi}{\ensuremath{+7.6}}

\newcommand{\cvHThreeNinetyLo}{\ensuremath{+2.0}}
\newcommand{\cvHThreeNinetyHi}{\ensuremath{+7.8}}
\newcommand{\cvHOneNinetyLo}{\ensuremath{-6.1}}
\newcommand{\cvHOneNinetyHi}{\ensuremath{-0.1}}
\newcommand{\cvKminusS}{\ensuremath{+1.7}}
\newcommand{\cvKminusSLo}{\ensuremath{-1.8}}
\newcommand{\cvKminusSHi}{\ensuremath{+5.2}}
\newcommand{\cvKminusSHard}{\ensuremath{-0.2}}
\newcommand{\cvHOneSeeds}{\ensuremath{-9}, \ensuremath{-11}, \ensuremath{-4}}
\newcommand{\cvHThreeSeeds}{\ensuremath{+7}, \ensuremath{+17}, \ensuremath{+13}}

%% file: macros_v7_auto.tex
\newcommand{\rvPool}{1{,}100}
\newcommand{\rvFail}{597}
\newcommand{\rvFailRate}{54}

\newcommand{\rvNtrain}{128}

\newcommand{\rvNseeds}{4}
\newcommand{\rvDerailStart}{566}
\newcommand{\rvDerailRel}{0.46}
\newcommand{\rvDerailKL}{1.49}
\newcommand{\rvControlKL}{0.32}

\newcommand{\rvRescueSixtyFour}{104}

\newcommand{\rvTeachLenSixtyFour}{84}

\newcommand{\rvRescueOneNineTwo}{111}

\newcommand{\rvRescueFiveTwelve}{164}

\newcommand{\rvTeachLenFiveTwelve}{530}

\newcommand{\rvBoxedFiveTwelve}{55}
\newcommand{\rvTeachEosFiveTwelve}{42}
\newcommand{\rvRescueS}{82}

\newcommand{\rvRescueN}{512}

\newcommand{\rvNestN}{512}

\newcommand{\rvGeoTokLong}{26.0\%}

\newcommand{\rvShareKL}{18\%}
\newcommand{\rvShareW}{52\%}
\newcommand{\rvShareLong}{45\%}
\newcommand{\rvAngleKL}{9}
\newcommand{\rvAngleW}{29}

\newcommand{\rvAngleLong}{24}

\newcommand{\rvDeltaAC}{\ensuremath{+4.9}}

\newcommand{\rvDeltaKL}{\ensuremath{+4.2}}

\newcommand{\rvDeltaW}{\ensuremath{+4.4}}

\newcommand{\rvDeltaLongAC}{\ensuremath{+4.5}}

\newcommand{\rvNsLongAC}{3}

\newcommand{\rvPOne}{\ensuremath{-0.6}}
\newcommand{\rvPOneLo}{\ensuremath{-2.6}}
\newcommand{\rvPOneHi}{\ensuremath{+1.3}}
\newcommand{\rvPOneP}{0.56}

\newcommand{\rvPTwo}{\ensuremath{+0.1}}
\newcommand{\rvPTwoLo}{\ensuremath{-1.9}}
\newcommand{\rvPTwoHi}{\ensuremath{+2.2}}
\newcommand{\rvPTwoP}{0.92}

\newcommand{\rvPOneFirst}{\ensuremath{+2.1}}
\newcommand{\rvPOneFirstLo}{\ensuremath{-0.8}}
\newcommand{\rvPOneFirstHi}{\ensuremath{+4.9}}

\newcommand{\rvPOneSecond}{\ensuremath{-3.3}}
\newcommand{\rvPOneSecondLo}{\ensuremath{-6.1}}
\newcommand{\rvPOneSecondHi}{\ensuremath{-0.7}}

\newcommand{\rvLongDelta}{\ensuremath{-2.0}}
\newcommand{\rvLongDeltaLo}{\ensuremath{-4.7}}
\newcommand{\rvLongDeltaHi}{\ensuremath{+0.8}}

\newcommand{\rvLongInter}{\ensuremath{-1.0}}

\newcommand{\rvLongSeeds}{2}
\newcommand{\rvFpPOne}{\ensuremath{-0.8}}

\newcommand{\rvFpPOneBb}{\ensuremath{-0.4}}

\newcommand{\rvSubFpCap}{76--92}

\newcommand{\rvSubFpCorrect}{32--49}

\newcommand{\rvSubInitFpCap}{9}
\newcommand{\rvSubInitFpCorrect}{48}
\newcommand{\rvSubBbCap}{11--18}

\newcommand{\rvSubBbCorrect}{60--70}

\newcommand{\rvSubInitBbCap}{4}
\newcommand{\rvSubInitBbCorrect}{60}

\newcommand{\rvSubInitOldFpCap}{5}
\newcommand{\rvSubInitOldBbCap}{11}
\newcommand{\rvSubUnits}{8}

\newcommand{\rvKLbf}{0.0038--0.0049}
\newcommand{\rvKLfp}{0.088--0.113}
\newcommand{\rvKLfold}{21--23}

\newcommand{\rvNormRatioPct}{7.6--8.2}
\newcommand{\rvDeltaCos}{0.67--0.68}
\newcommand{\rvKLnoise}{2.1\times10^{-4}}
\newcommand{\rvReverted}{93.4--93.9\%}

%% file: tables_auto_preprint.tex
\begin{table}[!htbp]\centering\small
\begin{adjustbox}{max width=\linewidth}\begin{tabular}{lrrrrrr}\toprule
Model & seed 1 & seed 2 & seed 3 & mean acc. (\%) & $\Delta$ vs initial (pp) & 95\% CI \\ \midrule
initial & \multicolumn{3}{c}{115/256} & 44.9 & -- & -- \\
allRKL & 129 & 126 & 128 & 49.9 & +4.9 & [+0.4, +9.4] \\
AC & 136 & 124 & 120 & 49.5 & +4.6 & [+0.3, +8.9] \\
mixedKL & 130 & 135 & 119 & 50.0 & +5.1 & [+0.5, +9.6] \\
randomK & 130 & 128 & 134 & 51.0 & +6.1 & [+1.6, +10.7] \\
pureS & 141 & 130 & 136 & 53.0 & +8.1 & [+3.0, +13.2] \\
noiseAC & \multicolumn{3}{c}{3 seeds} & 47.4 & +2.5 & [-2.0, +6.8] \\
\bottomrule\end{tabular}\end{adjustbox}
\caption{E1 (in-domain DeepMath, first 256 questions in frozen hash order): strict accuracy (explicit answer match and EOS). Seed columns are correct counts.}\label{tab:e1}
\end{table}
\begin{table}[!htbp]\centering\small
\begin{adjustbox}{max width=\linewidth}\begin{tabular}{lrrrrrr}\toprule
Model & seed 1 & seed 2 & seed 3 & mean acc. (\%) & $\Delta$ vs initial (pp) & 95\% CI \\ \midrule
initial & \multicolumn{3}{c}{224/256} & 87.5 & -- & -- \\
allRKL & 225 & 219 & 219 & 86.3 & -1.2 & [-4.6, +2.2] \\
AC & 221 & 222 & 223 & 86.7 & -0.8 & [-3.6, +2.2] \\
mixedKL & 224 & 219 & 217 & 85.9 & -1.6 & [-4.4, +1.2] \\
randomK & 218 & 224 & 219 & 86.1 & -1.4 & [-4.6, +1.6] \\
pureS & 221 & 220 & 219 & 85.9 & -1.6 & [-4.6, +1.4] \\
noiseAC & \multicolumn{3}{c}{3 seeds} & 86.5 & -1.0 & [-3.3, +1.2] \\
\bottomrule\end{tabular}\end{adjustbox}
\caption{E2 (GSM8K test subset, 256 questions, boxed-answer instruction): strict accuracy.}\label{tab:e2}
\end{table}
\begin{table}[!htbp]\centering\small
\begin{adjustbox}{max width=\linewidth}\begin{tabular}{lllrrrl}\toprule
Contrast & family step & TOST $\pm5$pp & E1 mean & E1 95\% CI & $p$ & E2 mean [95\% CI] \\ \midrule
AC-allRKL & 1: n.s., stop & yes [-3.1, +2.3] & -0.4 & [-3.6, +2.9] & 0.872 & +0.4 [-1.8, +2.7] \\
AC-mixedKL & 2: not tested & yes [-3.3, +2.2] & -0.5 & [-3.8, +2.7] & 0.819 & +0.8 [-1.2, +2.9] \\
mixedKL-allRKL & -- & yes [-2.6, +2.9] & +0.1 & [-3.3, +3.5] & 1.000 & -0.4 [-2.6, +1.8] \\
AC-randomK & 3: not tested & -- & -1.6 & [-4.8, +1.7] & 0.390 & +0.7 [-1.3, +2.7] \\
AC-pureS & 4: not tested & -- & -3.5 & [-7.0, -0.1] & 0.055 & +0.8 [-1.0, +2.6] \\
\bottomrule\end{tabular}\end{adjustbox}
\caption{Paired contrasts in percentage points (question-level bootstrap over the mean of 3 seeds, 10{,}000 resamples; $p$: sign-flip permutation). Family: pre-specified fixed sequence on E1; a contrast is tested only if every earlier one was significant. TOST: equivalent if the 90\% interval (shown) lies within $\pm5$\,pp. Rows below the first failed step are descriptive.}\label{tab:contrasts}
\end{table}
\begin{table}[!htbp]\centering\small
\begin{adjustbox}{max width=\linewidth}\begin{tabular}{lrrrr}\toprule
Arm & E1 $-$ noiseAC & E1 95\% CI & E2 $-$ noiseAC & E2 95\% CI \\ \midrule
allRKL & +2.5 & [-0.9, +5.7] & -0.1 & [-3.0, +2.6] \\
AC & +2.1 & [-1.2, +5.5] & +0.3 & [-2.3, +2.7] \\
mixedKL & +2.6 & [-0.8, +6.1] & -0.5 & [-2.9, +1.8] \\
randomK & +3.6 & [+0.4, +7.2] & -0.4 & [-3.1, +2.2] \\
pureS & +5.6 & [+2.2, +9.1] & -0.5 & [-3.3, +2.1] \\
noiseAC $-$ initial & +2.5 & [-2.0, +6.8] & -1.0 & [-3.3, +1.2] \\
\bottomrule\end{tabular}\end{adjustbox}
\caption{Trained arms against the matched random perturbation (noiseAC: $\theta_0$ plus a random-sign vector with the per-tensor norm of the same seed's AC update; mean over 3 seeds). Descriptive, outside the confirmatory family.}\label{tab:noise}
\end{table}

%% file: tables_v4_auto_preprint.tex
\begin{table}[!htbp]\centering\small
\begin{adjustbox}{max width=\linewidth}\begin{tabular}{llrrrrrr}\toprule
Dose & Arm & seed 1 / 2 / 3 & E1 acc. (\%) & $\Delta$ vs initial & 95\% CI & cap hits & E2 acc. (\%) \\ \midrule
0 & initial & \multicolumn{1}{c}{231/512} & 45.1 & -- & -- & -- & 87.5 \\
2 & AC & 235 / 226 / 243 & 45.8 & +0.7 & [-2.8, +4.2] & 24 & 86.6 \\
2 & pureS & 228 / 231 / 240 & 45.5 & +0.4 & [-3.1, +3.8] & 23 & 85.4 \\
\midrule
4 & AC & 258 / 256 / 248 & 49.6 & +4.5 & [+1.0, +7.9] & 41 & 85.5 \\
4 & pureS & 250 / 250 / 248 & 48.7 & +3.6 & [+0.3, +7.0] & 42 & 87.0 \\
\midrule
8 & AC & 230 / 248 / 235 & 46.4 & +1.3 & [-2.5, +5.1] & 178 & 86.3 \\
8 & pureS & 249 / 232 / 219 & 45.6 & +0.5 & [-3.6, +4.6] & 189 & 85.9 \\
\bottomrule\end{tabular}\end{adjustbox}
\caption{Confirmatory dose scan (protocol v4): E1 is the first 512 questions of the frozen order, E2 the same 256 GSM8K questions. Cap hits are E1 responses reaching 4096 tokens, mean over seeds. The initial student is the same sampled run as in Table~\ref{tab:e1}, now on 512 questions.}\label{tab:v4}
\end{table}
\begin{table}[!htbp]\centering\small\setlength{\tabcolsep}{3.5pt}
\begin{adjustbox}{max width=\linewidth}\begin{tabular}{llrrrrl}\toprule
Contrast & role & E1 mean & E1 95\% CI & E1 90\% CI & $p$ & E2 mean [95\% CI] \\ \midrule
AC$-$pureS, dose 4 & primary: n.s. & +0.9 & [-1.5, +3.2] & [-1.1, +2.9] & 0.476 & -1.4 [-3.6, +0.7] \\
AC$-$pureS, dose 2 & secondary: Holm $p$=1.00, n.s. & +0.3 & [-2.0, +2.7] & [-1.7, +2.3] & 0.825 & +1.2 [-0.9, +3.3] \\
AC$-$pureS, dose 8 & secondary: Holm $p$=1.00, n.s. & +0.8 & [-1.6, +3.3] & [-1.2, +2.9] & 0.529 & +0.4 [-1.7, +2.5] \\
\bottomrule\end{tabular}\end{adjustbox}
\caption{Pre-specified confirmatory contrasts (question-level bootstrap of the 3-seed mean, 10{,}000 resamples, seed 20261007; $p$: sign-flip permutation). Holm $p$-values include the monotonicity step; the frozen analysis script omitted it and printed 0.83 for dose 2. No decision changes.}\label{tab:v4contrast}
\end{table}

%% file: tables_cond_auto_preprint.tex
\begin{table}[!htbp]\centering\small\setlength{\tabcolsep}{3.5pt}
\begin{adjustbox}{max width=\linewidth}\begin{tabular}{lllcccrr}\toprule
Arm & Trajectory & Loss & Rescued & E1 correct & $\Delta$ vs.\ initial, pp & Cap & Med.\ len \\
 & & & (of 384) & seeds 1/2/3 & [95\% CI] & & \\ \midrule
initial & -- & -- & -- & 115 & -- & 11 & 1159 \\ \midrule
condT & $P_d\,|\,T\,|\,S$ & none on $T$ & 91 (23.7\%) & 121 / 119 / 129 & $\ensuremath{+3.1}$ $[\ensuremath{-1.6},\ensuremath{+7.7}]$ & 33 & 1459 \\
condS & $P_d\,|\,S$ & RKL, all & 66 (17.2\%) & 130 / 130 / 133 & $\ensuremath{+6.2}$ $[\ensuremath{+1.6},\ensuremath{+10.9}]$ & 25 & 1497 \\
randT & $P_c\,|\,T\,|\,S$ & none on $T$ & 71 (18.5\%) & 127 / 134 / 129 & $\ensuremath{+5.9}$ $[\ensuremath{+1.3},\ensuremath{+10.4}]$ & 29 & 1459 \\
condT$_{\mathrm{KL}}$ & $P_d\,|\,T\,|\,S$ & RKL, all & -- & 128 / 136 / 142 & $\ensuremath{+7.9}$ $[\ensuremath{+3.8},\ensuremath{+12.4}]$ & 27 & 1456 \\
trunc & $P_d$ & RKL, all & -- & 116 / 118 / 127 & $\ensuremath{+2.1}$ $[\ensuremath{-2.3},\ensuremath{+6.8}]$ & 13 & 1276 \\
\bottomrule\end{tabular}\end{adjustbox}
\caption{Conditional-intervention pilot (exploratory; E1, 256 questions). Each seed trains on its own 128 questions that the initial student had failed. $P_d$/$P_c$: the student's failed answer up to the derail/control point; $T$: teacher segment; $S$: student continuation; RKL: reverse KL. The derail point is the paragraph with the highest mean reverse KL, the control point a hashed paragraph at least two paragraphs away. Rescued: training trajectories whose final answer is strictly correct, before any training. Cap: mean number of E1 responses hitting the 4096-token cap; Med.\ len: mean over seeds of the median E1 response length. Intervals are question-level paired bootstraps of the seed mean.}\label{tab:cond}
\end{table}

\begin{table}[!htbp]\centering\small
\begin{adjustbox}{max width=\linewidth}\begin{tabular}{lccccc}\toprule
 & \multicolumn{3}{c}{E1 (256)} & \multicolumn{2}{c}{hard subset (141)} \\
Pre-specified contrast & pp & 95\% CI & wins/losses & pp & 95\% CI \\ \midrule
H1: condT $-$ condS (teacher as context) & $\ensuremath{-3.1}$ & $[\ensuremath{-6.6},\ensuremath{+0.4}]$ & 45/59 & $\ensuremath{-4.0}$ & $[\ensuremath{-9.0},\ensuremath{+0.9}]$ \\
H2: condT $-$ randT (placement) & $\ensuremath{-2.7}$ & $[\ensuremath{-6.1},\ensuremath{+0.7}]$ & 47/60 & $\ensuremath{-3.5}$ & $[\ensuremath{-8.7},\ensuremath{+1.7}]$ \\
H3: condT$_{\mathrm{KL}}$ $-$ condT (loss on teacher) & $\ensuremath{+4.8}$ & $[\ensuremath{+1.6},\ensuremath{+8.3}]$ & 57/37 & $\ensuremath{+3.8}$ & $[\ensuremath{-1.2},\ensuremath{+8.7}]$ \\
H4: condS $-$ trunc (continuation) & $\ensuremath{+4.2}$ & $[\ensuremath{+0.8},\ensuremath{+7.6}]$ & 60/39 & $\ensuremath{+4.7}$ & $[\ensuremath{0.0},\ensuremath{+9.7}]$ \\
\bottomrule\end{tabular}\end{adjustbox}
\caption{Pre-specified pilot contrasts (exploratory; no significance claims). Paired by question over the seed mean, 10{,}000 bootstrap draws. Wins/losses count questions on which the first arm's seed-mean score is higher/lower. The hard subset is the E1 questions the initial student answered wrongly, fixed before any training. The 90\% interval of H3 is $[\ensuremath{+2.0},\ensuremath{+7.8}]$\,pp: it excludes zero and is not inside the $\pm5$\,pp equivalence margin used in the main study.}\label{tab:condcontrast}
\end{table}

%% file: tables_v7_auto_preprint.tex
\begin{table}[!htbp]\centering\small\setlength{\tabcolsep}{3.5pt}
\begin{adjustbox}{max width=\linewidth}\begin{tabular}{lccccccc}\toprule
Arm & Teacher & $w$ & \shortstack{Teacher\\share} & Angle to & E1 correct & $\Delta$ vs.\ initial, pp & Cap \\
 & tokens & & of $\|g\|$ & AC gradient & seeds 1/2/3/4 & [95\% CI] & \\ \midrule
initial & -- & -- & -- & -- & 223 & -- & 17 \\ \midrule
T64$_{\mathrm{AC}}$ & 5.5\% & 0 & 0 & -- & 254 / 254 / 243 / 241 & $\ensuremath{+4.9}$ $[\ensuremath{+1.7},\ensuremath{+8.2}]$ & 52 \\
T64$_{\mathrm{KL}}$ & 5.5\% & 1 & 18\% & 9$^\circ$ & 245 / 253 / 228 / 253 & $\ensuremath{+4.2}$ $[\ensuremath{+0.9},\ensuremath{+7.6}]$ & 52 \\
T64$_{\mathrm{W4}}$ & 5.5\% & 4 & 52\% & 29$^\circ$ & 241 / 244 / 249 / 248 & $\ensuremath{+4.4}$ $[\ensuremath{+1.2},\ensuremath{+7.6}]$ & 41 \\
\midrule
T512$_{\mathrm{AC}}$ & 27.2\% & 0 & 0 & -- & 250 / 242 / 246 & $\ensuremath{+4.5}$ $[\ensuremath{+1.0},\ensuremath{+8.1}]$ & 53 \\
T512$_{\mathrm{KL}}$ & 26.0\% & 1 & 45\% & 24$^\circ$ & 236 / 236 & $\ensuremath{+2.5}$ $[\ensuremath{-1.1},\ensuremath{+6.2}]$ & 44 \\
\bottomrule\end{tabular}\end{adjustbox}
\caption{Pre-specified derail-point replication (E1, all 512 questions; 4 seeds of 128 failed training questions each). All arms train on $P_d\,|\,T\,|\,S$ trajectories with reverse KL on the student tokens; $w$ is the weight of the reverse KL on the teacher segment $T$ (0: context only). T64 is the pilot's teacher rule; T512 nests a longer prefix of the same teacher sample (length dose, secondary; its seed-3 KL unit was lost, Section~\ref{sec:deviations}). Teacher share of $\|g\|$: median of $w\|g_T\|/\|g_{\mathrm{stu}}+w\,g_T\|$ over all updates; angle: median angle between the arm's pre-clip gradient and the context-only gradient on the same batch. Cap: mean number of responses hitting the 4096-token cap. Intervals are question-level paired bootstraps of the seed mean.}\label{tab:v7}
\end{table}

\begin{table}[!htbp]\centering\small\setlength{\tabcolsep}{4pt}
\begin{adjustbox}{max width=\linewidth}\begin{tabular}{@{}>{\raggedright\arraybackslash}p{0.31\linewidth}rcrr>{\raggedright\arraybackslash}p{0.15\linewidth}@{}}\toprule
Contrast & pp & 95\% CI & $p$ & wins/losses & \\ \midrule
P1: T64$_{\mathrm{KL}}-$T64$_{\mathrm{AC}}$ (primary) & $\ensuremath{-0.6}$ & $[\ensuremath{-2.6},\ensuremath{+1.3}]$ & 0.56 & 105/109 & Holm $\alpha=0.025$, not confirmed \\
P2: T64$_{\mathrm{W4}}-$T64$_{\mathrm{KL}}$ (primary) & $\ensuremath{+0.1}$ & $[\ensuremath{-1.9},\ensuremath{+2.2}]$ & 0.92 & 108/100 & Holm $\alpha=0.050$, not confirmed \\
\midrule
P1 on E1 1--256 (pilot's set) & $\ensuremath{+2.1}$ & $[\ensuremath{-0.8},\ensuremath{+4.9}]$ & 0.17 & 62/45 &  \\
P1 on E1 257--512 (never in the pilot) & $\ensuremath{-3.3}$ & $[\ensuremath{-6.1},\ensuremath{-0.7}]$ & 0.02 & 43/64 &  \\
P2 on E1 1--256 & $\ensuremath{-0.1}$ & $[\ensuremath{-2.9},\ensuremath{+2.7}]$ & 1.00 & 53/51 &  \\
P2 on E1 257--512 & $\ensuremath{+0.4}$ & $[\ensuremath{-2.4},\ensuremath{+3.2}]$ & 0.84 & 55/49 &  \\
T64$_{\mathrm{W4}}-$T64$_{\mathrm{AC}}$ & $\ensuremath{-0.5}$ & $[\ensuremath{-2.5},\ensuremath{+1.5}]$ & 0.66 & 109/106 &  \\
T512$_{\mathrm{KL}}-$T512$_{\mathrm{AC}}$ & $\ensuremath{-2.0}$ & $[\ensuremath{-4.7},\ensuremath{+0.8}]$ & 0.18 & 64/88 & 2 seeds \\
length interaction (T512 $-$ T64) & $\ensuremath{-1.0}$ & $[\ensuremath{-5.0},\ensuremath{+2.9}]$ & 0.66 & 108/107 & 2 seeds \\
P1 on E1 1--128, exact FP32 & $\ensuremath{-0.8}$ & $[\ensuremath{-4.3},\ensuremath{+2.5}]$ & 0.75 & 19/24 & batch 32 \\
P1 on E1 1--128, BF16 autocast & $\ensuremath{-0.4}$ & $[\ensuremath{-4.9},\ensuremath{+4.1}]$ & 0.93 & 29/30 & batch 32 \\
\bottomrule\end{tabular}\end{adjustbox}
\caption{Contrasts of the replication. Paired by question over the seed mean; 10{,}000 bootstrap draws; $p$ is a two-sided sign-flip permutation test. P1 and P2 are the pre-specified family (Holm at 0.05; the 97.5\% interval of P1 is $[\ensuremath{-2.8},\ensuremath{+1.6}]$); every other row is secondary. Wins/losses count questions on which the first arm's seed-mean score is higher/lower.}\label{tab:v7contrast}
\end{table}

\begin{table}[!htbp]\centering\small
\begin{adjustbox}{max width=\linewidth}\begin{tabular}{lccccc}\toprule
 & \multicolumn{3}{c}{KL(trained $\|$ initial) per token} & \multicolumn{2}{c}{\shortstack{log-prob change $\delta$\\on realized tokens}} \\
Arm & BF16 autocast & exact FP32 & ratio & $\frac{\|\delta_{\mathrm{BF16}}\|}{\|\delta_{\mathrm{FP32}}\|}$ & $\cos(\delta_{\mathrm{BF16}},\delta_{\mathrm{FP32}})$ \\ \midrule
T64$_{\mathrm{AC}}$ & 0.0049 & 0.113 & 23$\times$ & 0.076 & 0.67 \\
T64$_{\mathrm{KL}}$ & 0.0047 & 0.109 & 23$\times$ & 0.076 & 0.67 \\
T64$_{\mathrm{W4}}$ & 0.0038 & 0.088 & 23$\times$ & 0.082 & 0.67 \\
T512$_{\mathrm{AC}}$ & 0.0048 & 0.111 & 23$\times$ & 0.080 & 0.68 \\
T512$_{\mathrm{KL}}$ & 0.0047 & 0.101 & 21$\times$ & 0.080 & 0.68 \\
\bottomrule\end{tabular}\end{adjustbox}
\caption{Function-space probe (medians over seeds). Each trained model and the initial student score the same 64 fixed responses (the initial student's E1 answers), once under the BF16 autocast used by every evaluation in this paper and once in exact FP32 (no autocast, no TF32, math attention); full-vocabulary KL is normalized in FP64. The numerical noise floor, KL between the initial student under the two numerics, is $2.1\times10^{-4}$.}\label{tab:v7num}
\end{table}

\begin{table}[!htbp]\centering\small
\begin{adjustbox}{max width=\linewidth}\begin{tabular}{lccccc}\toprule
 & \multicolumn{2}{c}{initial student} & \multicolumn{3}{c}{trained (T64$_{\mathrm{AC}}$, T64$_{\mathrm{KL}}$; 8 models)} \\
Evaluation of E1 1--128 & correct & cap & correct & cap & med.\ length \\ \midrule
BF16 autocast, batch 64 & 58 & 6 & 66.5 (61--72) & 15.5 (14--19) & 1363 \\
BF16 autocast, batch 32 & 60 & 4 & 65.5 (60--70) & 14.1 (11--18) & 1397 \\
exact FP32, batch 32 & 48 & 9 & 41.0 (32--49) & 82.9 (76--92) & 4096 \\
\bottomrule\end{tabular}\end{adjustbox}
\caption{The same models evaluated under two numerics on the first 128 E1 questions (same common random numbers). Trained columns give the mean over models and the range. Every trained model hit the 4096-token cap more often, and answered fewer questions correctly, in exact FP32 than in BF16 at the same batch size. The initial student hit the cap 9 times in FP32 (5 on the main study's random stream).}\label{tab:v7fp}
\end{table}

%% file: tables_v5_auto_preprint.tex
\begin{table}[!htbp]\centering\small
\begin{adjustbox}{max width=\linewidth}\begin{tabular}{lllcccl}\toprule
 & & & step 25 & step 50 & step 100 & \\
Recipe & lr & Change & dev128 & dev128 & dev256 & Outcome \\ \midrule
initial & -- & -- & \multicolumn{2}{c}{62 / 4} & 120 / 8 & -- \\ \midrule
R1 & $10^{-6}$ & EOS mask & 64 / 31 & -- & -- & aborted at 25 \\
R2 & $3\times10^{-7}$ & EOS mask & 61 / 9 & 60 / 14 & -- & aborted at 50 \\
R3 & $10^{-6}$ & -- & 61 / 31 & -- & -- & aborted at 25 \\
R4 & $3\times10^{-7}$ & -- & 59 / 6 & 60 / 11 & 132 / 59 & completed; failed: cap \\
R5 & $10^{-6}$ & EOS mask + anchor & 64 / 0 & 65 / 27 & -- & aborted at 50 \\
\midrule
R7 & $3\times10^{-7}$ & ref.\ KL, $\beta=0.5$ & 61 / 6 & 61 / 13 & -- & aborted at 50 \\
R6 & $3\times10^{-7}$ & teacher Qwen3-4B & 57 / 6 & 56 / 9 & 119 / 21 & completed; failed: cap, accuracy \\
\bottomrule\end{tabular}\end{adjustbox}
\caption{Online OPD recipe search (exploratory; AC objective, 16 fresh student rollouts per step, 100 steps). Cells are strict correct / responses hitting the 4096-token cap. Anchor: $-\log p_\theta(\mathrm{EOS})$ at final states, $\lambda=1$; Qwen3-4B in non-thinking mode. Steps 25 and 50 use the first 128 dev questions, step 100 all 256. The pre-specified abort rule stopped a recipe when dev128 cap hits exceeded 12 (three times the initial). The gate required at most 16 cap hits and at least 128 correct on dev256, and no decline after step 50. R5 was added by amendment 1, R6 and R7 by amendment 3, each before it ran.}\label{tab:v5}
\end{table}

%% file: appendix_auto_preprint.tex
\begingroup
\setlength{\intextsep}{5pt}
\begin{table}[H]\centering\small
\begin{adjustbox}{max width=\linewidth}\begin{tabular}{@{}p{0.30\linewidth}p{\dimexpr0.70\linewidth-2\tabcolsep\relax}@{}}\toprule
Student / teacher & Qwen3-1.7B (FP32 master weights) / Qwen3-4B-Instruct-2507 (BF16, frozen) \\
Training questions & 128 DeepMath training questions, identical for every seed \\
Trajectory caps & prefix 256 (cut at first paragraph), teacher paragraph $\le$1024 (cut at first paragraph), total 4096 \\
Sampling for trajectories & T=1, top-$p$=1, top-$k$=0; EOS ids 151643, 151645 \\
Updates / batch & 4 / 32 (one pass over 128 trajectories) \\
Optimizer & AdamW, lr $3\times10^{-6}$, $\beta=(0.9,0.95)$, $\epsilon=10^{-8}$, no weight decay, clip 1.0 \\
Loss & full-vocabulary token KL, $\sum_t w_t D_t / L$ per sequence, mean over batch; chunk 128; BF16 autocast forward \\
Seeds (rollout / permutation / mask) & seed1: 606301/606311/606321; seed2: 606302/606312/606322; seed3: 606303/606313/606323 \\
Evaluation decoding & sample: T=0.7, top-$k$=20, top-$p$=0.8; cap 4096; batch 64; common random numbers per question \\
E1 / E2 & first 256 of e1\_ordered512 (DeepMath, hash order) / first 256 of e2\_ordered512 (GSM8K test) \\
Online seeds & rollout 606401+step, permutation 606411, mask 606421 \\
Config & config\_s3\_v3\_1.json, sha256 65c08d7d\ldots; statistics: 10{,}000 question-level bootstrap resamples, seed 20261006 \\
\bottomrule\end{tabular}\end{adjustbox}
\caption{Frozen configuration of the final run.}\label{tab:config}
\end{table}
\begin{table}[H]\centering\small
\begin{adjustbox}{max width=\linewidth}\begin{tabular}{lrrrrrrr}\toprule
Unit group & $n$ upd. & B tok. share & B KL share & $\|g_B\|/\|g_{AC}\|$ & $\cos(g_B^{rev},g_{AC})$ & $\cos(g_{AC},g_{AC}{+}g_B)$ & pre-clip norm \\ \midrule
S3 allRKL & 12 & 2.7\% & 1.8\% & 0.080 & 0.126 & 0.997 & 5.50 \\
S3 AC & 12 & 2.7\% & 1.8\% & 0.080 & 0.128 & 0.997 & 5.27 \\
S3 mixedKL & 12 & 2.7\% & 1.8\% & 0.080 & 0.126 & 0.997 & 5.27 \\
S3 randomK & 12 & 2.7\% & 1.8\% & 0.080 & 0.125 & 0.997 & 5.35 \\
S4 online (3 arms) & 12 & 2.9\% & 1.7\% & 0.077 & 0.095 & 0.997 & 5.40 \\
\bottomrule\end{tabular}\end{adjustbox}
\caption{Mechanism medians by unit group: S3 is the fixed-trajectory main study, and S4 the fresh-rollout on-policy check. Each S3 group pools 3 seeds $\times$ 4 updates. The quantities are measured at each arm's own parameters, so the groups agree only because the arms stay close.}\label{tab:mech}
\end{table}
\begin{table}[H]\centering\footnotesize\setlength{\tabcolsep}{2.5pt}
\begin{adjustbox}{max width=\linewidth}\begin{tabular}{llrrrrrrrrrr}\toprule
Seed & Arm & \shortstack{E1\\corr.} & \shortstack{E1\\EOS} & \shortstack{E1\\unparsed} & \shortstack{E1\\tokens} & \shortstack{E2\\corr.} & \shortstack{E2\\EOS} & $\|\Delta\theta\|_2$ & \shortstack{BF16\\cos} & \shortstack{BF16\\reverted} & \shortstack{train\\s} \\ \midrule
1 & allRKL & 129 & 229 & 42 & 1805 & 225 & 256 & 0.304 & 0.30 & 93.5\% & 338 \\
1 & AC & 136 & 227 & 44 & 1789 & 221 & 256 & 0.305 & 0.30 & 93.5\% & 159 \\
1 & mixedKL & 130 & 229 & 46 & 1808 & 224 & 256 & 0.305 & 0.30 & 93.5\% & 164 \\
1 & randomK & 130 & 234 & 41 & 1869 & 218 & 256 & 0.305 & 0.30 & 93.5\% & 196 \\
1 & pureS & 141 & 229 & 42 & 1827 & 221 & 256 & 0.303 & 0.30 & 93.6\% & 52 \\
1 & noiseAC & 121 & 246 & 38 & 1395 & 220 & 256 & 0.305 & 0.30 & 91.8\% & -- \\
\midrule
2 & allRKL & 126 & 234 & 40 & 1780 & 219 & 256 & 0.304 & 0.30 & 93.6\% & 161 \\
2 & AC & 124 & 227 & 43 & 1807 & 222 & 256 & 0.303 & 0.30 & 93.6\% & 163 \\
2 & mixedKL & 135 & 228 & 45 & 1802 & 219 & 256 & 0.303 & 0.30 & 93.6\% & 163 \\
2 & randomK & 128 & 235 & 45 & 1739 & 224 & 256 & 0.303 & 0.30 & 93.6\% & 197 \\
2 & pureS & 130 & 228 & 47 & 1822 & 220 & 256 & 0.307 & 0.30 & 93.5\% & 52 \\
2 & noiseAC & 119 & 242 & 37 & 1504 & 221 & 256 & 0.303 & 0.29 & 91.9\% & -- \\
\midrule
3 & allRKL & 128 & 237 & 39 & 1715 & 219 & 255 & 0.303 & 0.30 & 93.6\% & 158 \\
3 & AC & 120 & 230 & 48 & 1813 & 223 & 256 & 0.306 & 0.30 & 93.5\% & 157 \\
3 & mixedKL & 119 & 231 & 48 & 1842 & 217 & 256 & 0.305 & 0.30 & 93.6\% & 158 \\
3 & randomK & 134 & 230 & 44 & 1800 & 219 & 256 & 0.302 & 0.30 & 93.6\% & 194 \\
3 & pureS & 136 & 239 & 38 & 1764 & 219 & 256 & 0.305 & 0.30 & 93.6\% & 52 \\
3 & noiseAC & 124 & 248 & 40 & 1396 & 223 & 256 & 0.306 & 0.30 & 91.7\% & -- \\
\bottomrule\end{tabular}\end{adjustbox}
\caption{Per-unit results on the common 256 E1 questions and 256 E2 questions. $\|\Delta\theta\|_2$ is the FP32 parameter change from $\theta_0$; BF16 cos is the cosine between that change and its BF16-rounded image; BF16 reverted is the fraction of changed elements that round back to $\theta_0$. Training seconds include GPU sharing with concurrent collection for the first unit.}\label{tab:units}
\end{table}
\begin{table}[H]\centering\small
\begin{adjustbox}{max width=\linewidth}\begin{tabular}{lrrrrrrr}\toprule
Arm (on-policy, 1 seed) & E1 corr. & E1 EOS & E1 tokens & E2 corr. & $\|\Delta\theta\|_2$ & BF16 cos & gen. tokens/step \\ \midrule
allRKL & 127 & 236 & 1765 & 222 & 0.306 & 0.30 & 84147 \\
AC & 117 & 229 & 1865 & 218 & 0.306 & 0.30 & 82084 \\
mixedKL & 121 & 227 & 1865 & 217 & 0.305 & 0.30 & 81563 \\
\bottomrule\end{tabular}\end{adjustbox}
\caption{Fresh-rollout on-policy check (S4; 4 updates of 32 freshly sampled trajectories each). E1 is the common 256-question set.}\label{tab:online}
\end{table}
\begin{table}[H]\centering\small
\begin{adjustbox}{max width=\linewidth}\begin{tabular}{llrrrr}\toprule
Model & decoding & cap hits /64 & EOS /64 & mean tokens & correct /64 \\ \midrule
initial & sampling & 3 & 61 & 1319 & 24 \\
allRKL, 16 upd. & sampling & 43 & 21 & 3533 & 15 \\
allRKL, 16 upd. & greedy & 42 & 22 & 3384 & 16 \\
AC, 16 upd. & sampling & 42 & 22 & 3361 & 17 \\
mixedKL, 16 upd. & sampling & 38 & 26 & 3416 & 19 \\
pureS, 16 upd. & sampling & 46 & 18 & 3535 & 15 \\
\bottomrule\end{tabular}\end{adjustbox}
\caption{Termination at the original dose (16 updates of 8 trajectories, seed-1 trajectories) on 64 previously used historical questions. Sampling = the recommended Qwen3 settings with common random numbers. Correct counts are descriptive only.}\label{tab:collapse}
\end{table}

\endgroup